\documentclass[conference]{IEEEtran}
\usepackage{times}

\usepackage[numbers]{natbib}
\usepackage{multicol}
\usepackage[bookmarks=true]{hyperref}

\usepackage{tikz}
\usetikzlibrary{shapes.geometric, arrows}

\usepackage{multirow}

\usepackage{textcomp}

\usepackage{subcaption}

\usepackage{graphicx}

\begin{document}

\title{Active Control of Camera Parameters for Object Detection Algorithms}


\author{\authorblockN{Yulong Wu, John Tsotsos}
\authorblockA{Department of Electrical Engineering and Computer Science\\
York Univeristy\\
Toronto, ON M3J 1P3\\
Email: \{yulong, tsotsos\}@cse.yorku.ca}
}

\maketitle

\begin{abstract}
Camera parameters not only play an important role in determining the visual quality of perceived images, but also affect the performance of vision algorithms, for a vision-guided robot. By quantitatively evaluating four object detection algorithms, with respect to varying ambient illumination, shutter speed and voltage gain, it is observed that the performance of the algorithms is highly dependent on these variables. From this observation, a novel active control of camera parameters method is proposed, to make robot vision more robust under different light conditions. Experimental results demonstrate the effectiveness of our proposed approach,  which improves the performance of object detection algorithms, compared with the conventional auto-exposure algorithm.
\end{abstract}

\IEEEpeerreviewmaketitle

\section{Introduction}
\label{introduction}

In the 1980s, \citet{bajcsy1985active} introduced the concept of active perception, as ``a problem of intelligent control strategies applied to the data acquisition process". This idea was later explored and termed ``active vision", with an emphasis on visual perception, by \citet{aloimonos1988active}. In their studies, it was shown that many vision problems could be solved in a much more efficient way by an active approach than a passive one. Active vision was later formalized as a special case of the attention problem, by Tsotsos \cite{tsotsos1992relative}.

Despite the advantages of being active, most vision-guided robotic systems are characterized by their passive perspectives. Most of them rely on camera's built-in auto-exposure algorithms \cite{johnson1984photographic, sampat1999system}, which set camera exposure by evaluating the mean brightness of an image. While these methods result in \textit{good} images from the perspective of human, it is not always the case for a robot. Moreover, vision algorithms are typically trained on offline image datasets, which suffer from a significant camera sensor specific bias \cite{andreopoulos2012sensor}. This results in less generalized models, which are sensitive to camera parameters and often fail on poor exposed images.

\begin{figure}[h]
	\centering
	\includegraphics[trim=2.3cm 7.5cm 2.8cm 7cm,clip,width=0.48\textwidth]{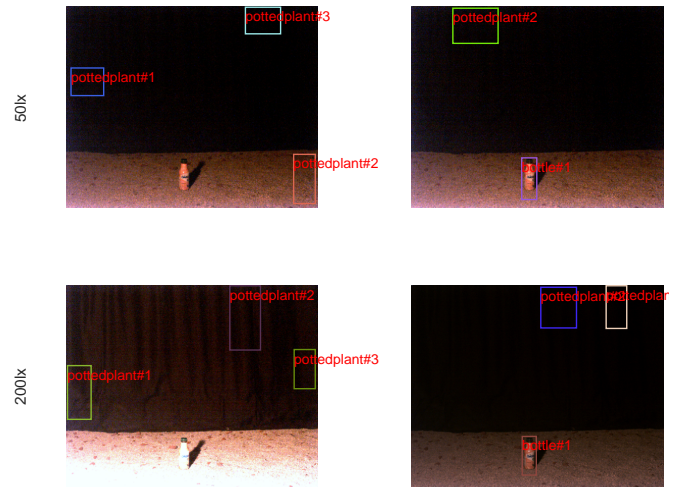}
	\caption{The top-3 outputs of the deformable parts model algorithm on the auto-exposure images (left column) and best-performing images (right column; images with the camera parameters that yielded the best object detection results). The first and second rows are for 50lx and 200lx illumination respectively.}
	\label{fig_intro_example}
\end{figure}

Figure~\ref{fig_intro_example} demonstrates a failure case of an object detection algorithm when using auto-exposure. For this case, the illumination is low, 50lx and 200lx. The built-in auto-exposure uses very large shutter speed and voltage gain to compensate for this low light condition.  While it increases the overall brightness of the acquired image, it fails the object detection algorithm as large shutter speed results in over-exposure around the object of interest and large voltage gain  introduces noise. This example indicates that a finer control of camera's intrinsic parameters is needed.

In this paper, we present a novel active control of camera parameters method, to make robot vision more robust against variation in illumination. Specifically, we investigate object detection algorithms, as object detection is one of the basic tasks in vision-guided robots. For camera parameters, we focus on shutter speed and voltage gain. There are manly two contributions in this work: 1) quantitative evaluation of object detection algorithms reveals their sensitivity to camera parameters ; 2) a novel active control of camera parameters method is proposed to improve the robustness of vision algorithms.

\section{Related Work}
\label{related_work}

How to make robot vision robust to different light conditions is still a challenging problem in the robotics community \cite{adini1997face, osadchy2001image, osadchy2004efficient, tanaka2009object, maier2011image, linderoth2013color}, as a slight ambient illumination change may produce large difference in the  appearance of objects. In the literature, there are four common approaches to achieve this goal, from different perspectives.

The first one is to use illumination-insensitive representations of an image, such as edge maps \cite{wei2004robust}, features in the frequency domain derived for a differentiated image \cite{llano2006illumination} and inferred albedo and surface normal from neural networks \cite{TangSH12b}. Better illumination invariance could be achieved by using these representations instead of the original image. The second approach is to use multiple instance-based models, where each instance corresponds to one light condition. Belhumeur \cite{belhumeur1998set} proved that the set of images of an object in fixed pose but with variant illumination, forms a convex cone, and the dimension of this illumination cone equals the number of distinct surface normals. However, algorithms based on this approach typically need large amount of training data and have high computational cost. The third approach is camera sensor accommodation \cite{tenenbaum1970accommodation}, which dates back to the 1970s.  It was proposed that sensor accommodation, automatic control by computer over the parameters of camera, should be an integral part of the recognition process. This idea was later applied on active fixation in the context of object recognition \cite{brunnstrom1996active}.  The fourth one is illumination preprocessing. Preprocessing has been a common practice in object recognition pipelines, which aims to improve the reliability of a vision system. For face recognition, particularly, a study \cite{han2013comparative} demonstrated that illumination preprocessing is helpful in handling lighting variations.

In this paper, we use the third approach to achieve the robustness of object detection algorithms to different light conditions. It was proposed that camera parameters should be optimized with respect to different metrics, like image entropy in \cite{lu2010camera} and gradient information in \cite{shim2014auto}. This methodology was further summarized in \cite{westerhoff2015generic}. In comparison with the previous work, our method emphasizes that the control of camera parameters should be optimized by the performance of specific vision applications, i.e. object detection algorithms, rather than using a generic metric.

\section{Camera Parameters on the Performance of Object Detection Algorithms}
\label{cam_params_on_object_detection}

As introduced in Section \ref{introduction},  camera parameters determine the quality of perceived images, which affect the performance of vision algorithms. In this section, we present our quantitative evaluation of object detection algorithms, with respect to different ambient illuminations and two camera settings, i.e. shutter speed and voltage gain.

\subsection{Dataset}
\label{dataset}

One of the fundamental problems of common image datasets, such as \cite{bileschi2007cbcl, everingham2010pascal, russakovsky2015imagenet, lin2014microsoft}, is that they are image sensor biased. Most images are taken under good lighting condition and with proper exposure.  In order to evaluate the sensitivity of object detection algorithms to camera parameters, a new dataset is introduced in this paper.

\begin{figure}[h]
	\centering
	\includegraphics[width=0.4\textwidth]{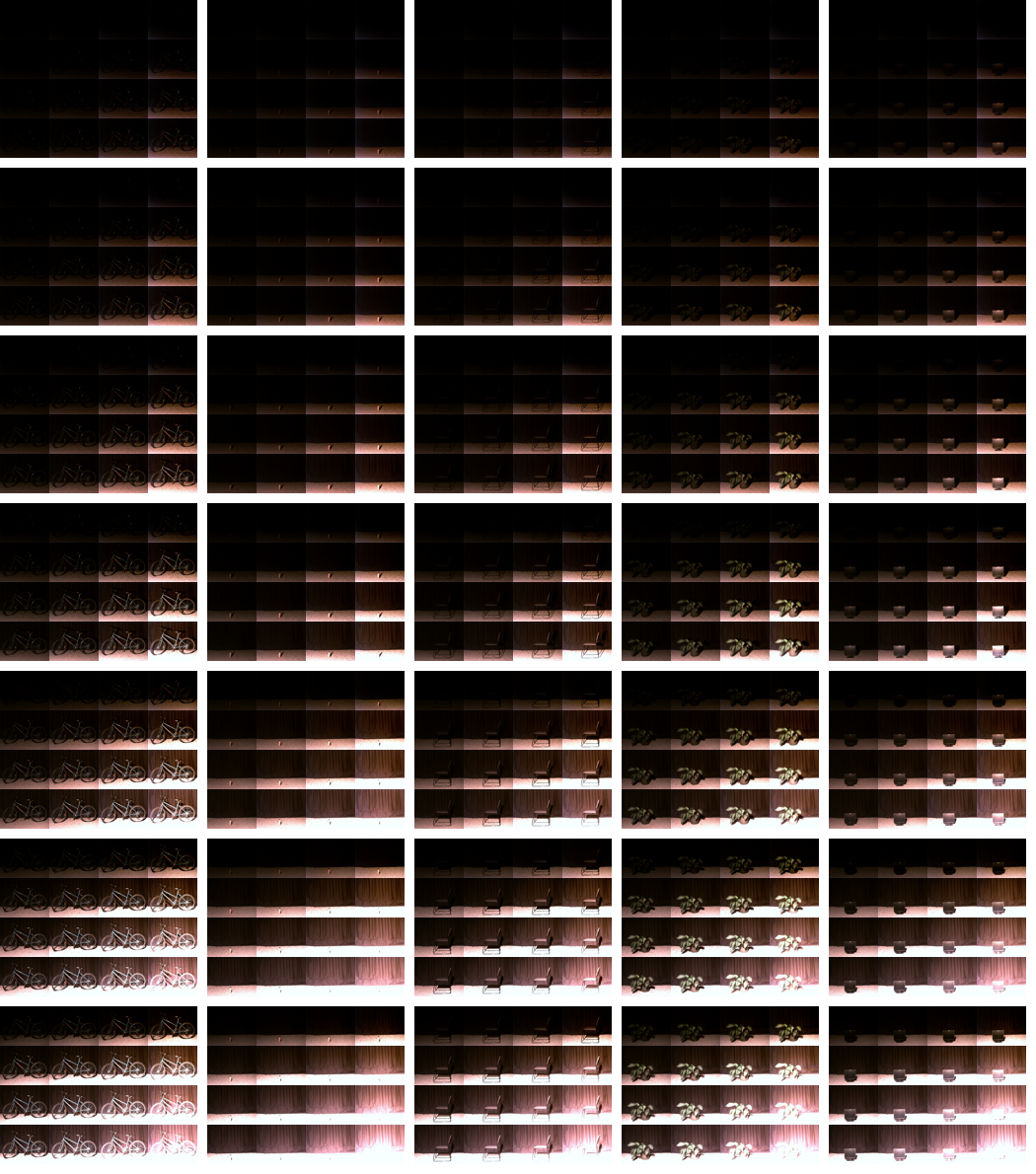}
	\caption{Sample images from the dataset. Each row corresponds to one light condition, 50lx to 3200lx from top to bottom. Each column corresponds to one object. For each illumination and object, there are 64 images (only 16 of them are displayed), by sampling the shutter speed and voltage gain parameters. Best viewed in high-resolution display.}
	\label{fig_dataset}
\end{figure}

This dataset contains 2240 images in total, by viewing 5 different objects (\textit{bicycle}, \textit{bottle}, \textit{chair}, \textit{pottedplant} and \textit{tvmonitor}), at 7 levels of illumination and with 64 camera configurations (8 shutter speeds $\times$ 8 voltage gains). Each image is in 8-bit/color RGB format and of 1280x1204 resolution. All object instances in the dataset are manually annotated with class labels and  bounding boxes. Samples of this dataset can be found in Figure~\ref{fig_dataset}. The full dataset is published at \url{http://jtl.lassonde.yorku.ca/software/datasets/}.

To accurately measure the illumination of the scene, a Yoctopuce light sensor \footnote{http://www.yoctopuce.com/EN/products/usb-environmental-sensors/yocto-light-v3} was used to measure the ambient illumination. Also, intensity-controllable light bulbs were used to achieve different light conditions, 50lx, 200lx, 400lx, 800lx, 1600lx and 3200lx. The digital camera was a Point Grey Flea3 camera (mode: FL3-U3-13E4C-C), which was equipped with a CMOS sensor and a programmable API interface. The allowed shutter speed and voltage gain ranges were 0.016ms-24.973ms and 0dB-24.014dB respectively. These permissible ranges were uniformly sampled into 8 distinct values in each dimension. The $i^{th}$ sample from a range $[a, b]$ was set as $a + \frac{b - a}{8} (i - 1)$, where $i \in \{1, ..., 8\}$, leading to $8\times8$ candidate settings for the shutter/gain parameters, under which the corresponding images were acquired.  The aperture was fixed at 4, and the red and blue white-balancing channels were set to 500 and 800 respectively. All other parameters were kept at default values.

\subsection{Evaluation Setup}
From the literature, four popular object detection algorithms were selected for evaluating: the Deformable Part Models (DPM) \cite{felzenszwalb2010object}, the Bag-of-Words Model with Spatial Pyramid Matching (BoW) \cite{uijlings2013selective}, the Regions with Convolutional Neural Networks (R-CNN) \cite{girshick2014rich}, and the Spatial Pyramid Pooling in Deep Convolutional Networks (SPP-net) \cite{he2015spatial}. The original implementations were used (except for BoW), and no optimization or transfer learning techniques were applied. 

For the DPM, the Release 5 version, as published at \url{https://people.eecs.berkeley.edu/~rbg/latent/}, was adopted. There were twenty class-specific detectors trained on the PASCAL VOC 2007 dataset \cite{everingham2010pascal}, and only five of them were used for the purpose of this evaluation. The outputs of each detector were combined using non-maximum suppression with a 0.5 overlap threshold. For the BoW, we replicated the idea in \cite{uijlings2013selective} with our own implementation. To make it consistent with the other algorithms, it was trained only on the PASCAL VOC 2007 dataset. Local features were sampled densely over the images and represented by SIFT \cite{lowe2004distinctive} and HoG \cite{dalal2005histograms} descriptors. We used a visual book size of 1000 and a spatial pyramid with 3 levels using 1x1, 2x2, and 4x4. For the classifiers, we used linear SVMs with a chi-squared kernel. The retraining process took two iterations. For the R-CNN and SPP-net, the neural networks were pre-trained on ImageNet and fine-tuned on the PASCAL VOC 2007 dataset. Twenty detectors were trained for the objects in the PASCAL dataset, and only five of them were used for evaluating. The outputs of each detector were combined via non-maximum suppression with an overlap threshold of 0.5.

The output, given an input image, of each algorithm was required to be a list of predicted object instances, each represented by a bounding box, a level and a confidence score. A predicted instance is considered true if the label is correct and the bounding box overlaps no less than 50\% with the ground-truth bounding box, otherwise false.

Following the methodology by Andreopoulos \& Tsotsos \cite{andreopoulos2012sensor}, the evaluation procedures include:
\begin{enumerate}
	\item Run the object detection algorithms on all the images that correspond to each $\langle illumination, shutter, gain\rangle$ combination;
	\item Sort the outputs by their confidence scores and then evaluate them, using the aforementioned  rule;
	\item Compute the precision-recall curve from the above results;
	\item Compute the average precision (AP) by sampling the precision-recall curve.
\end{enumerate}

The final results are represented by performance tables. A performance table is a 8x8 matrix $M$, where $M_{ij}$ is the AP of an algorithm on all the images that correspond to $i^{th}$ sample of shutter speed and $j^{th}$ sample of voltage gain, for a illumination. The range of AP is $[0, 1]$. Larger APs are represented in black color, and smaller are in white color.

\subsection{Results and Discussion}

\begin{figure}[h]
	\centering
	
	\begin{subfigure}[b]{0.22\textwidth}
		\includegraphics[draft=false,width=\textwidth]{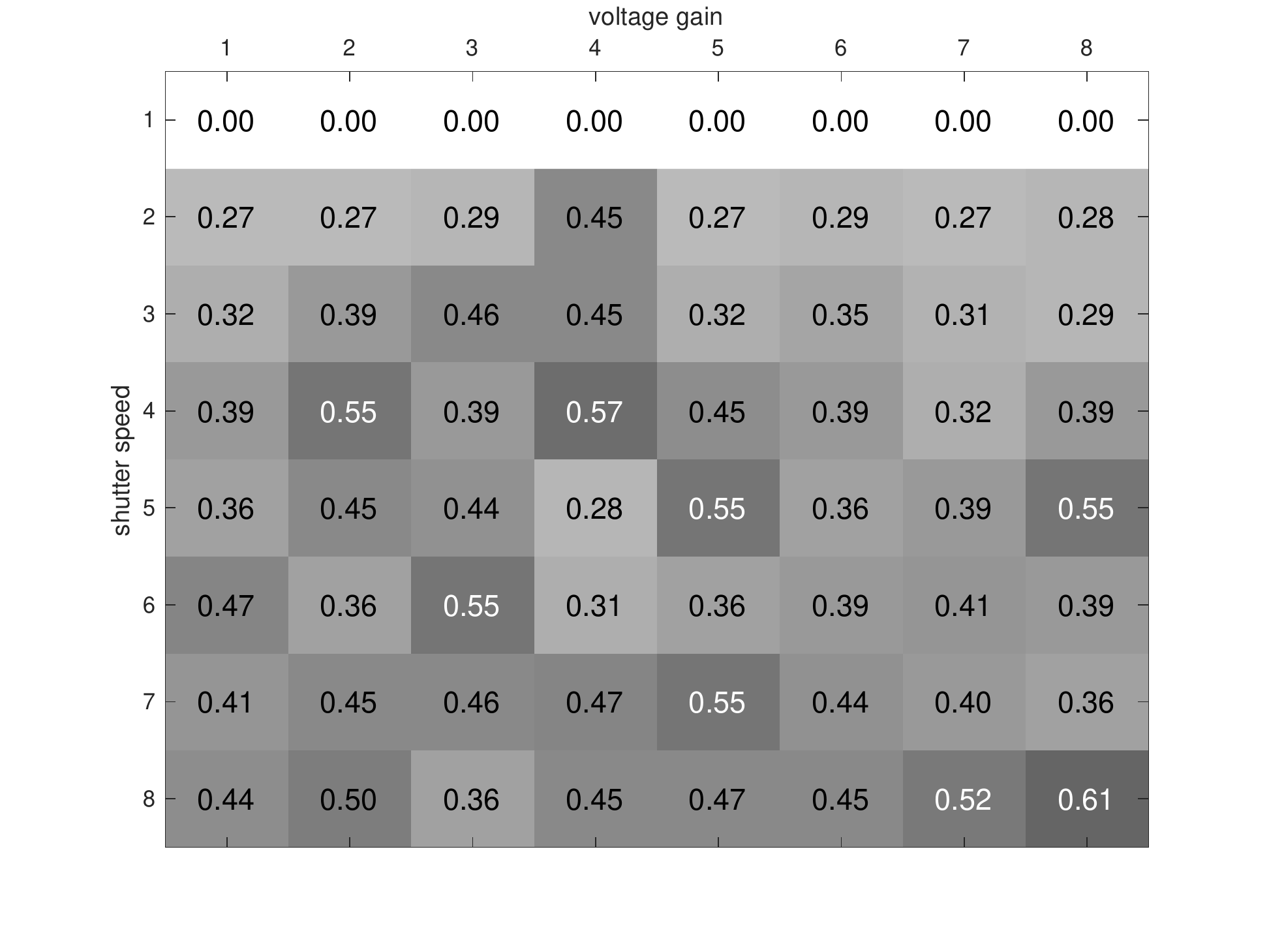}
		\caption{DPM}
	\end{subfigure}
	~
	\begin{subfigure}[b]{0.22\textwidth}
		\includegraphics[draft=false,width=\textwidth]{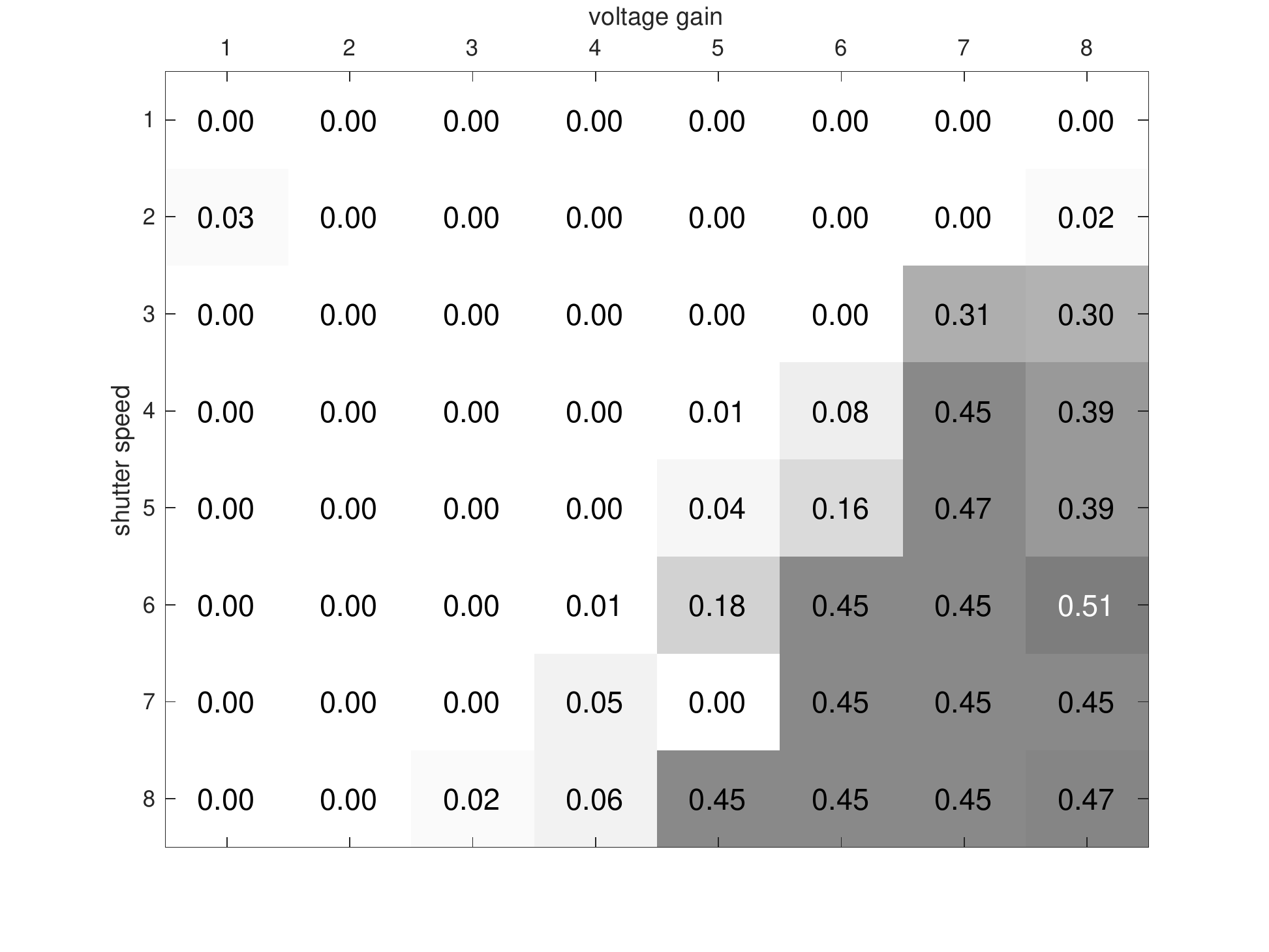}
		\caption{BoW}
	\end{subfigure}
	
	\begin{subfigure}[b]{0.22\textwidth}
		\includegraphics[draft=false,width=\textwidth]{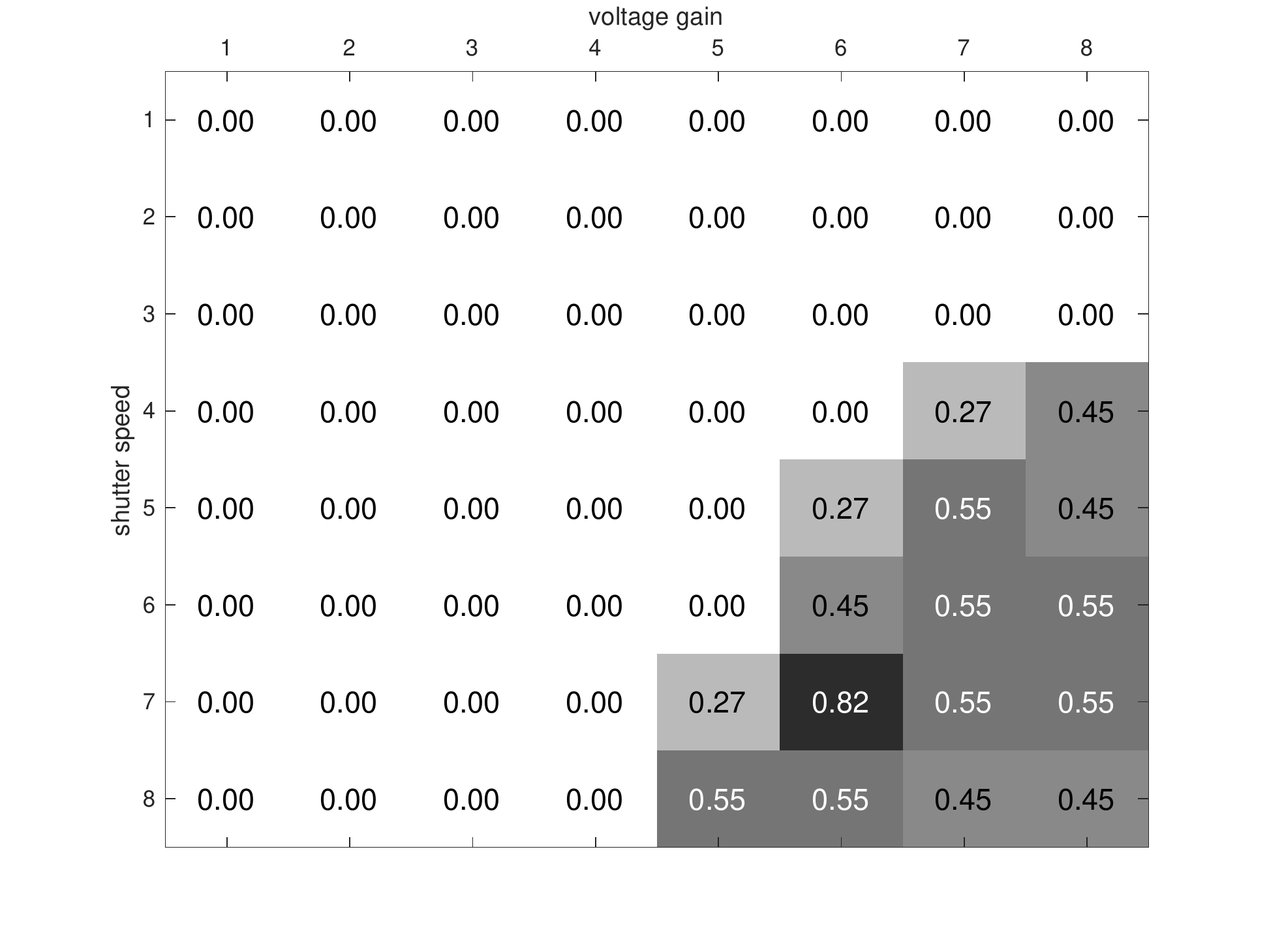}
		\caption{RCNN}
	\end{subfigure}
	~
	\begin{subfigure}[b]{0.22\textwidth}
		\includegraphics[draft=false,width=\textwidth]{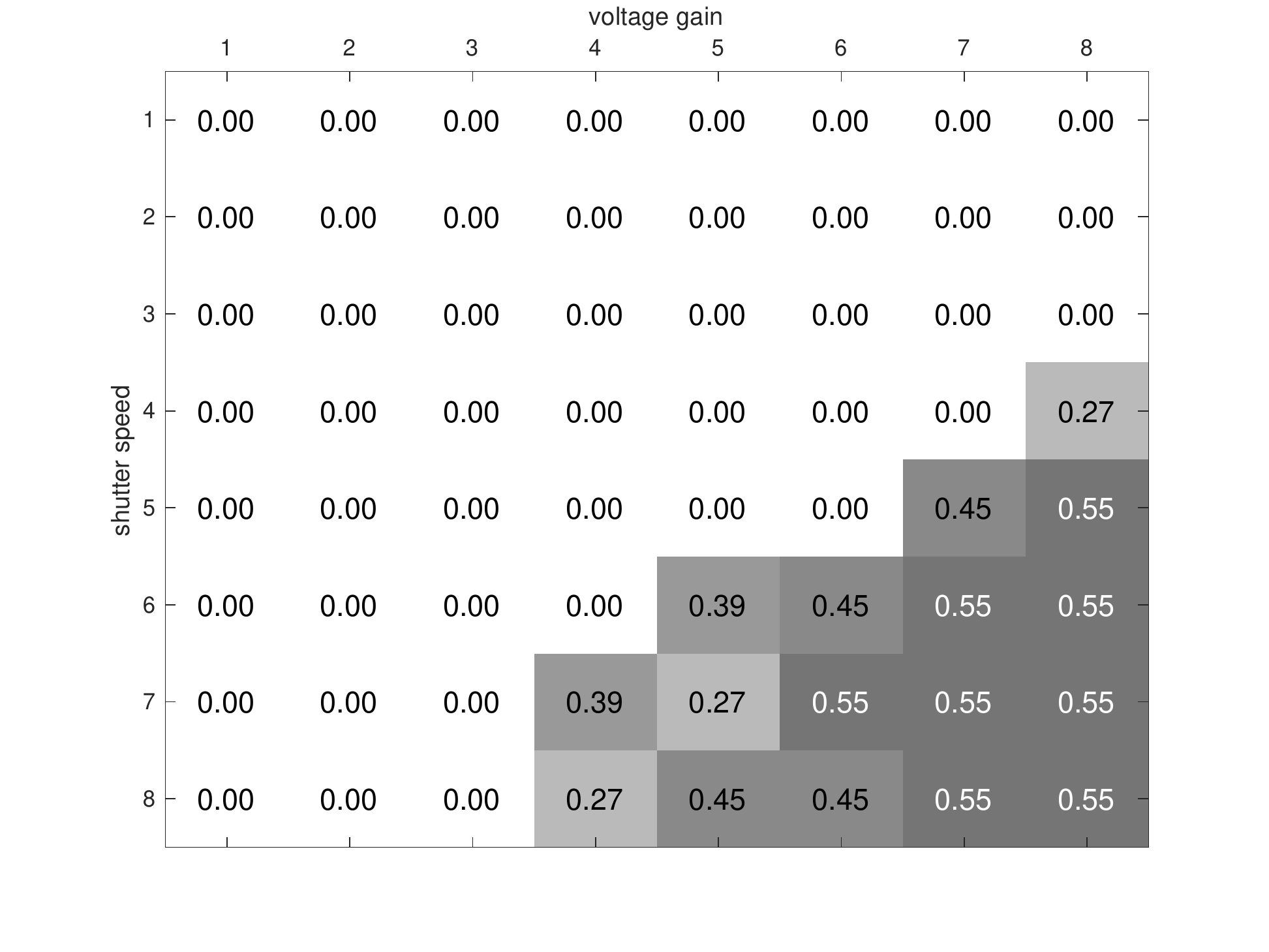}
		\caption{SPP-net}
	\end{subfigure}
	
	\caption{The performance of four algorithms on the images taken under low illumination (50lx).  Inside each performance table, the shutter speed increases from top to bottom and the voltage gain increases from left to right.}
	\label{fig_evaluation_results_1}
\end{figure}

\begin{figure}[h]
	\centering
	
	\begin{subfigure}[b]{0.22\textwidth}
		\includegraphics[draft=false,width=\textwidth]{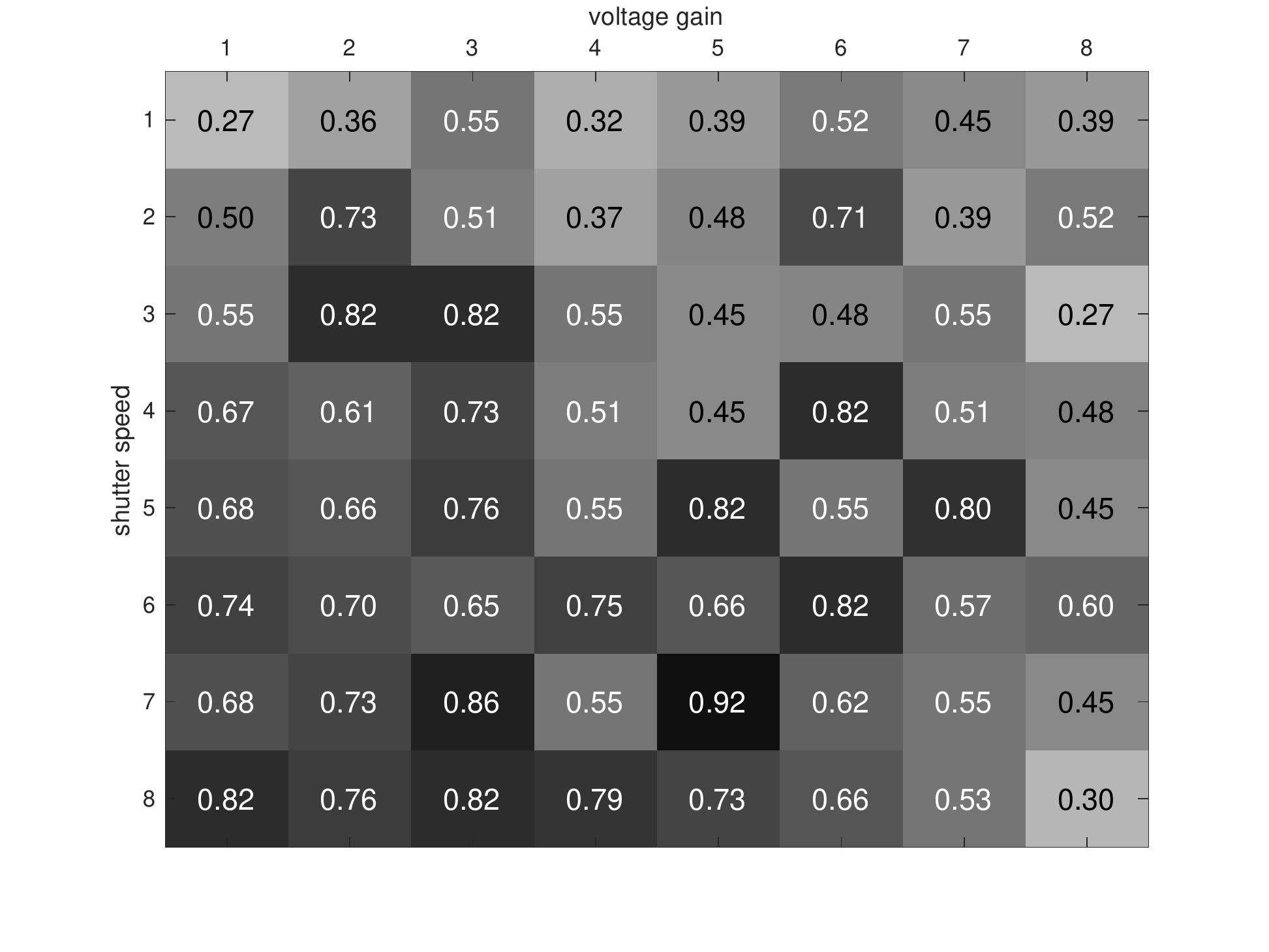}
		\caption{DPM}
	\end{subfigure}
	~
	\begin{subfigure}[b]{0.22\textwidth}
		\includegraphics[draft=false,width=\textwidth]{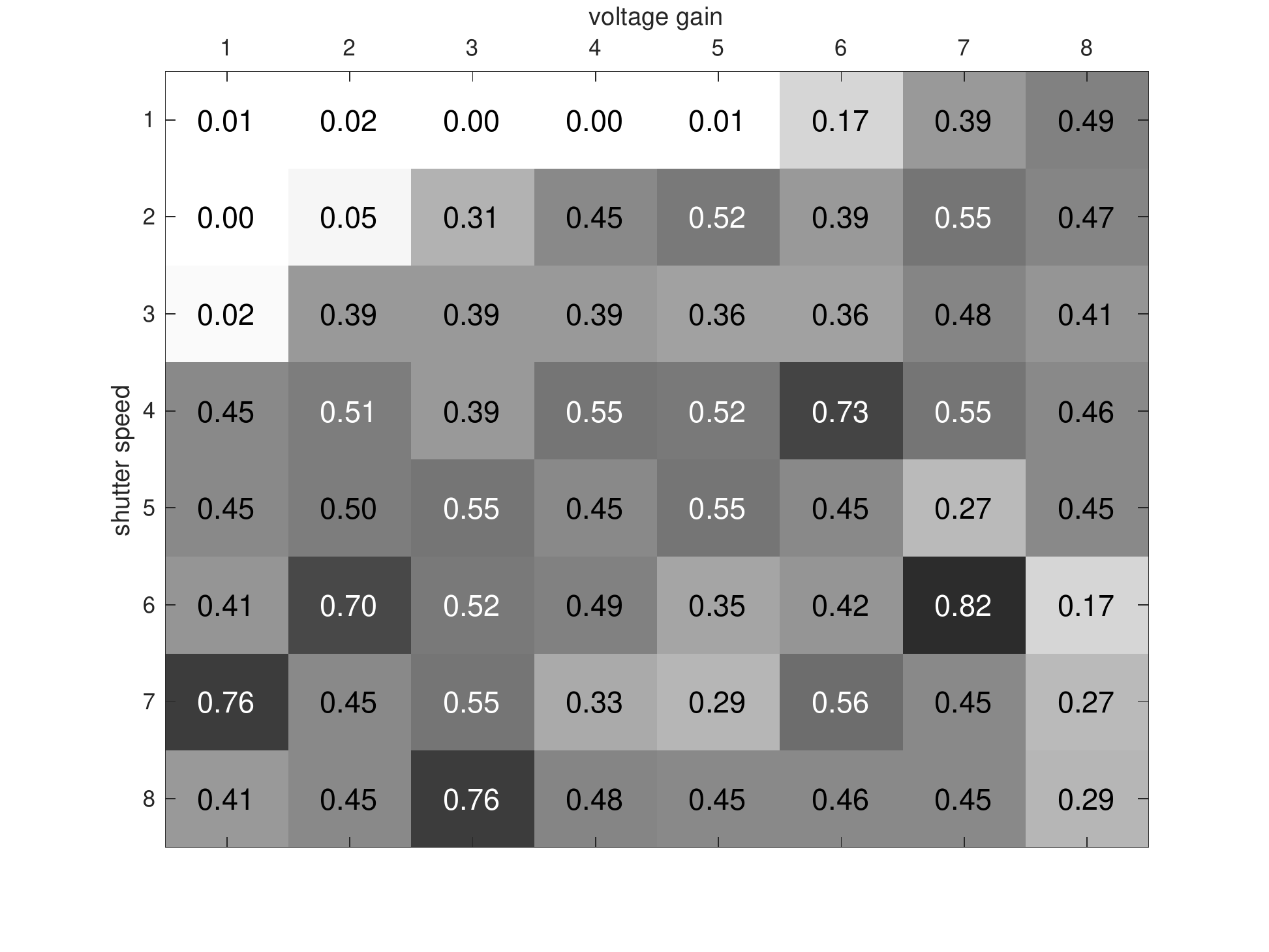}
		\caption{BoW}
	\end{subfigure}
	
	\begin{subfigure}[b]{0.22\textwidth}
		\includegraphics[draft=false,width=\textwidth]{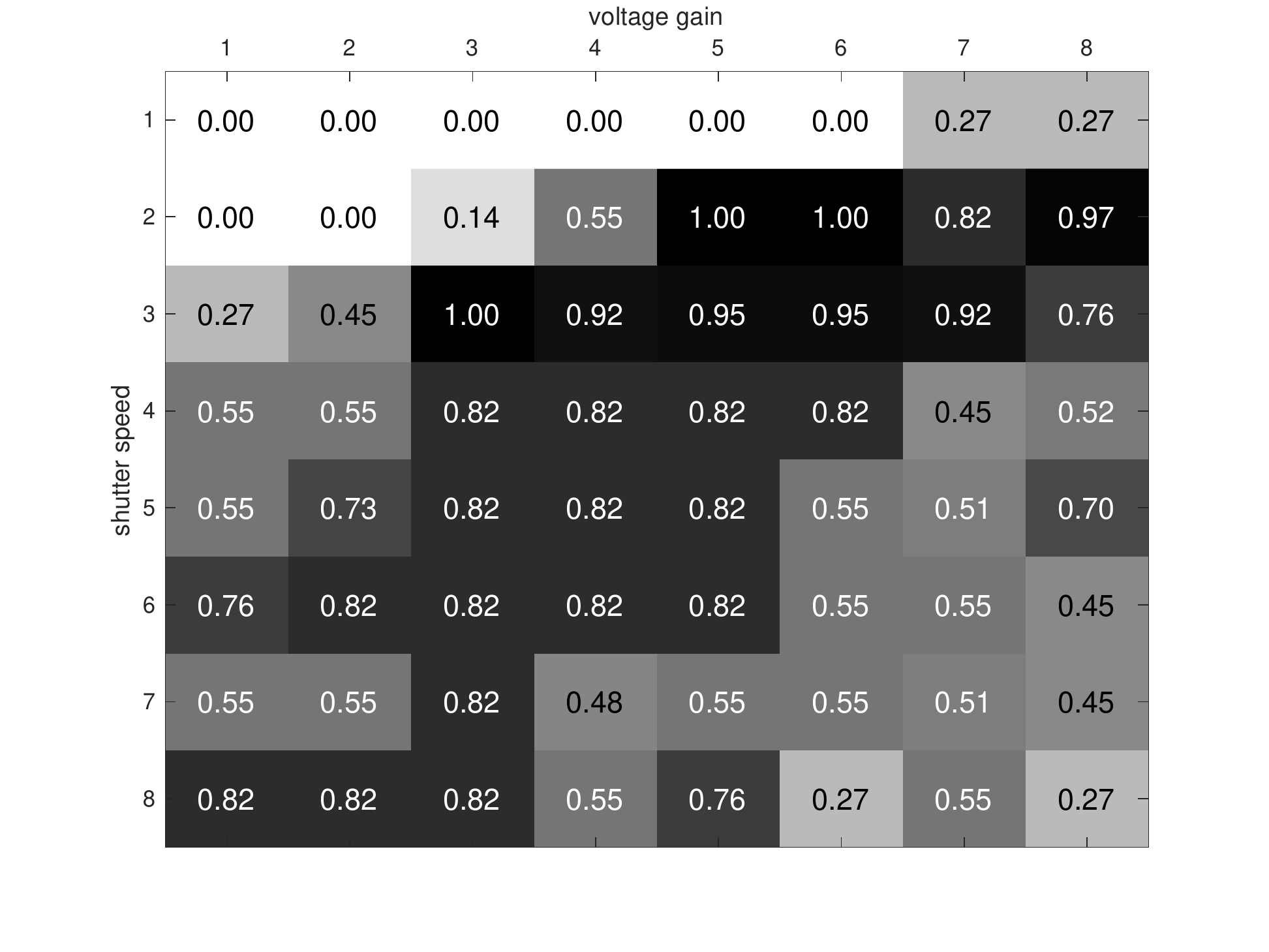}
		\caption{RCNN}
	\end{subfigure}
	~
	\begin{subfigure}[b]{0.22\textwidth}
		\includegraphics[draft=false,width=\textwidth]{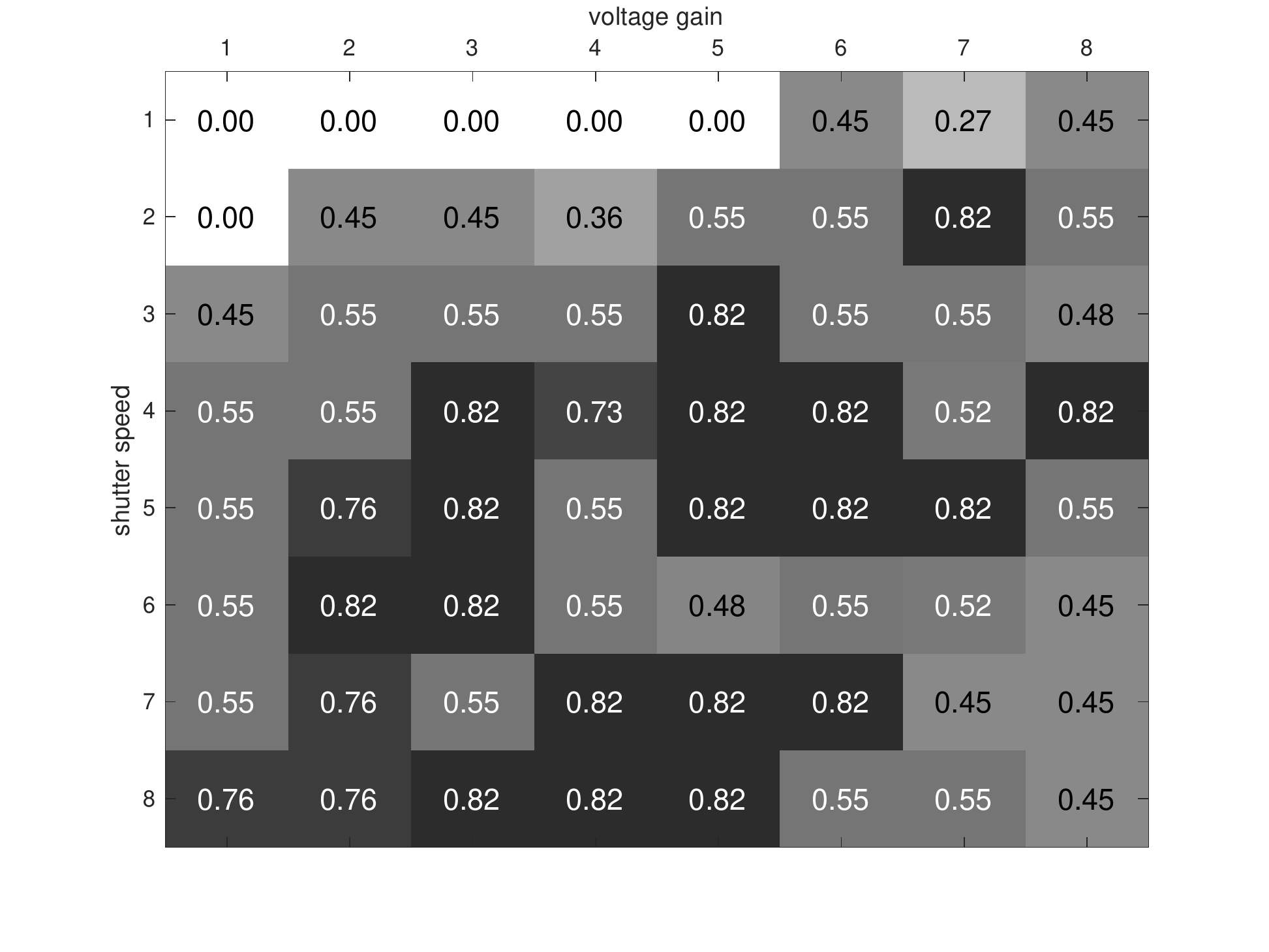}
		\caption{SPP-net}
	\end{subfigure}
	
	\caption{The performance of four algorithms on the images taken under medium illumination (400lx).  Inside each performance table, the shutter speed increases from top to bottom and the voltage gain increases from left to right.}
	\label{fig_evaluation_results_2}
\end{figure}

\begin{figure}[h]
	\centering
	
	\begin{subfigure}[b]{0.22\textwidth}
		\includegraphics[draft=false,width=\textwidth]{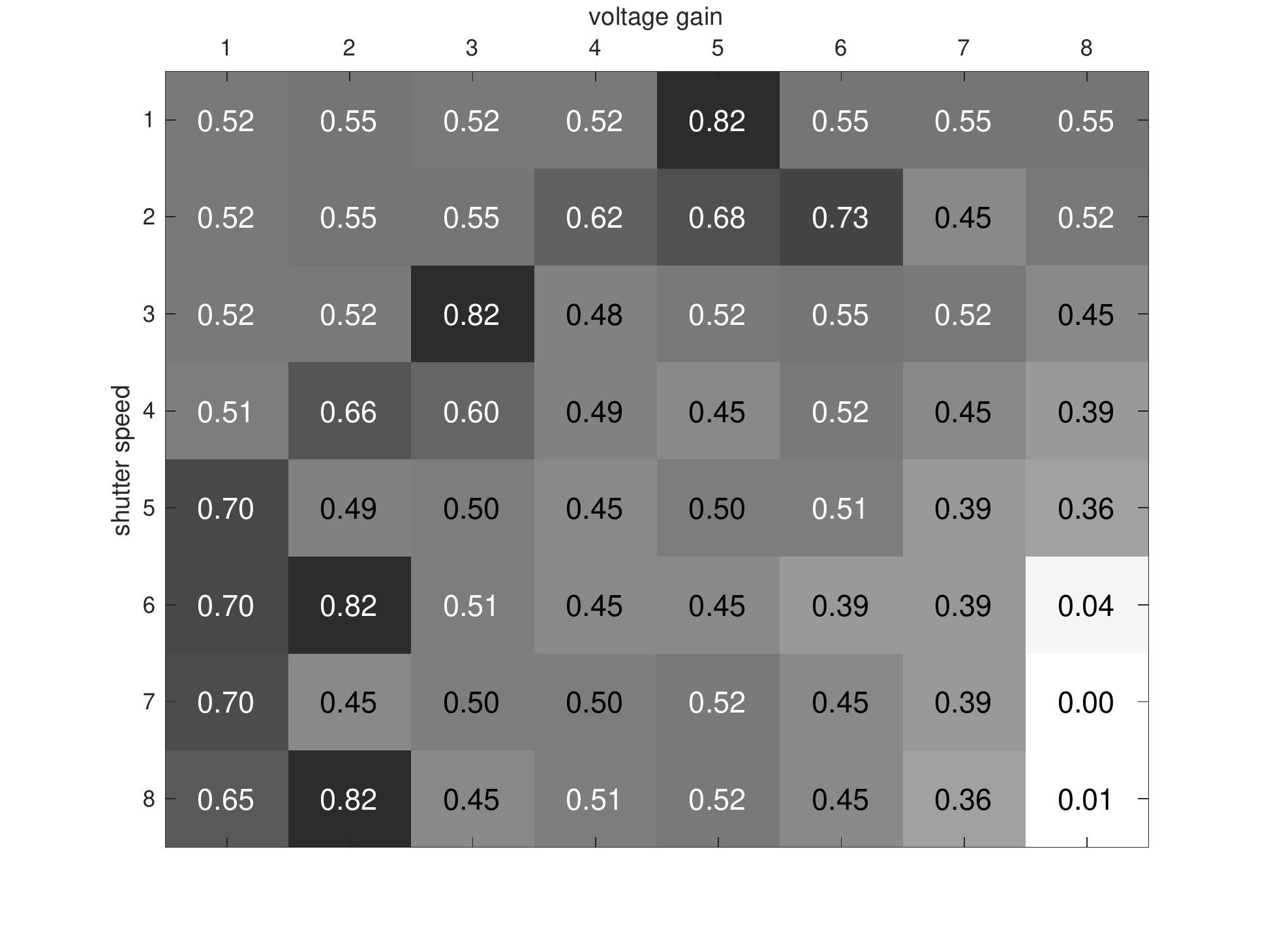}
		\caption{DPM}
	\end{subfigure}
	~
	\begin{subfigure}[b]{0.22\textwidth}
		\includegraphics[draft=false,width=\textwidth]{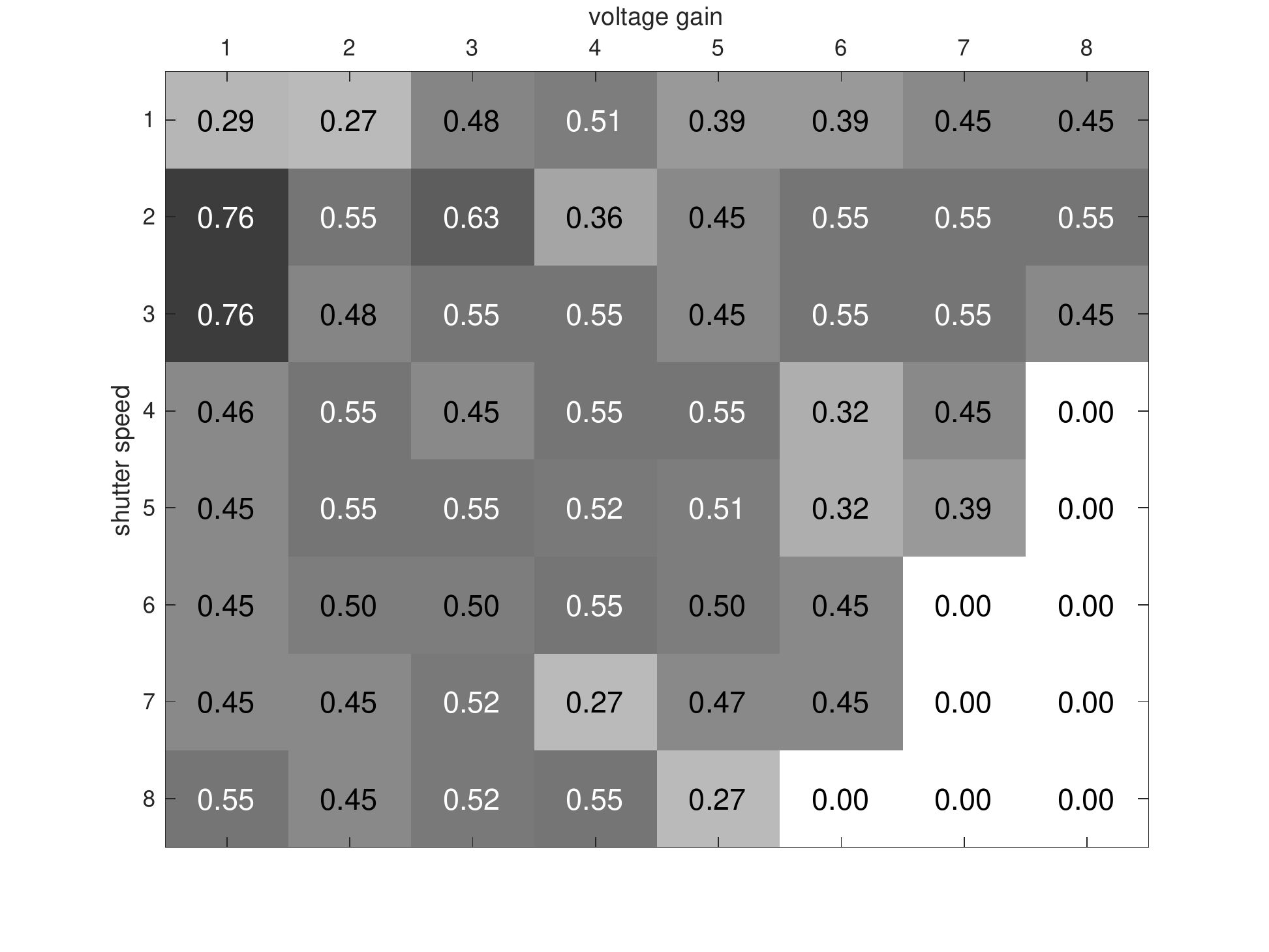}
		\caption{BoW}
	\end{subfigure}
	
	\begin{subfigure}[b]{0.22\textwidth}
		\includegraphics[draft=false,width=\textwidth]{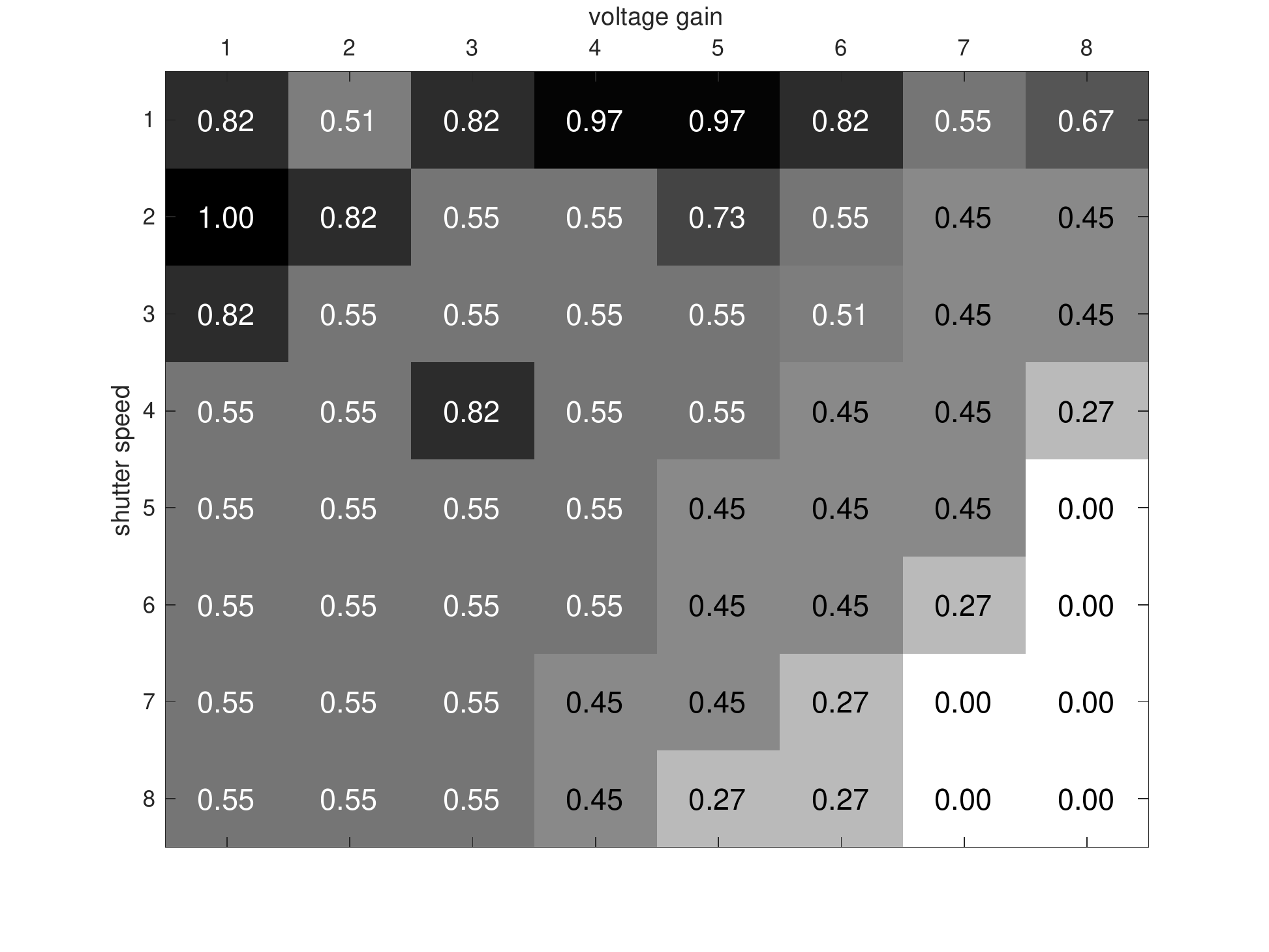}
		\caption{RCNN}
	\end{subfigure}
	~
	\begin{subfigure}[b]{0.22\textwidth}
		\includegraphics[draft=false,width=\textwidth]{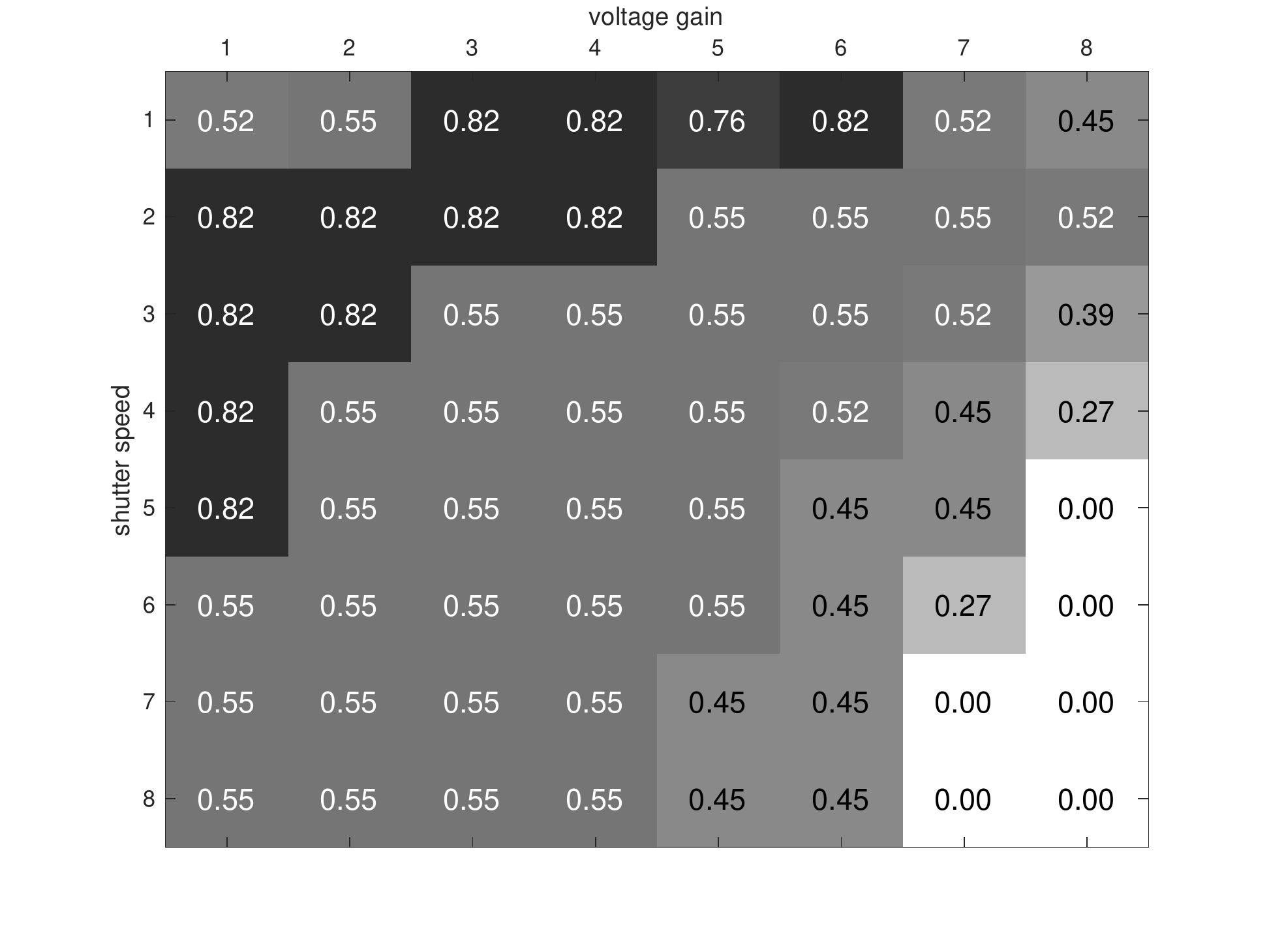}
		\caption{SPP-net}
	\end{subfigure}
	
	\caption{The performance of four algorithms on the images taken under high illumination (3200lx).  Inside each performance table, the shutter speed increases from top to bottom and the voltage gain increases from left to right.}
	\label{fig_evaluation_results_3}
\end{figure}

Figure \ref{fig_evaluation_results_1} - \ref{fig_evaluation_results_3} shows the performance tables of object detection algorithms under three light conditions. The most obvious observation is that all algorithms only work with a subset of the $\langle shutter, gain \rangle$ pairs, for a specific illumination. Another observation is that the algorithms prefer faster shutter speed and smaller voltage gain when the scene is bright, and slower shutter speed and larger voltage gain when the scene is dark. However, algorithms demonstrate different sensitivity to changes in shutter speed and voltage gain. The DPM accepts wider range of values in the shutter/gain parameter space due to the relative illumination robustness of the underlying HOG features, while the BoW, R-CNN and SPP-net work with narrower range of values. Also, the best-performing camera parameters, for each illumination, vary among algorithms.

\begin{figure}[h]
	\centering
	
	\begin{subfigure}[b]{0.23\textwidth}
		\includegraphics[draft=false,width=\textwidth]{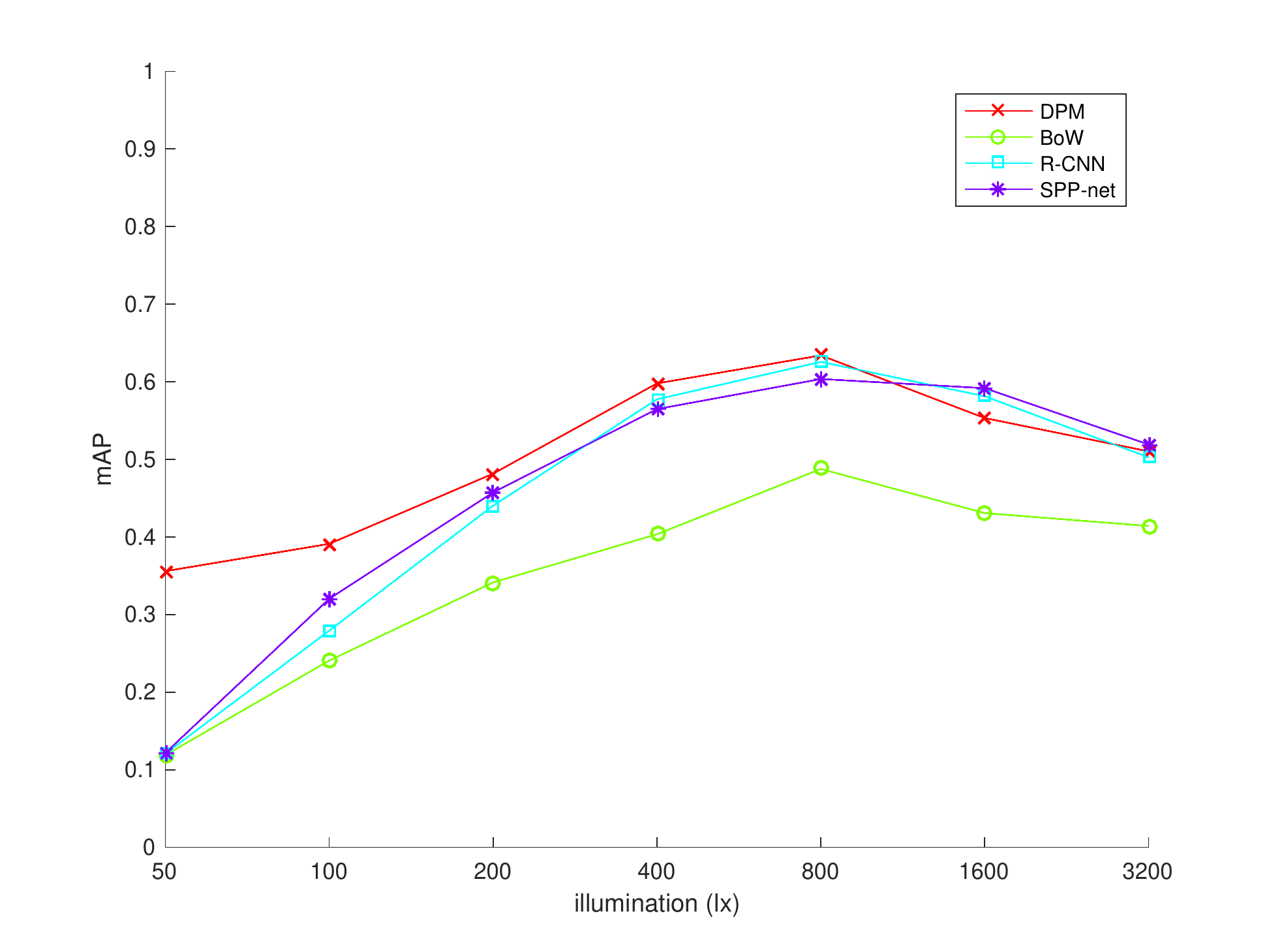}
		\caption{Illumination}
	\end{subfigure}
	~
	\begin{subfigure}[b]{0.23\textwidth}
		\includegraphics[draft=false,width=\textwidth]{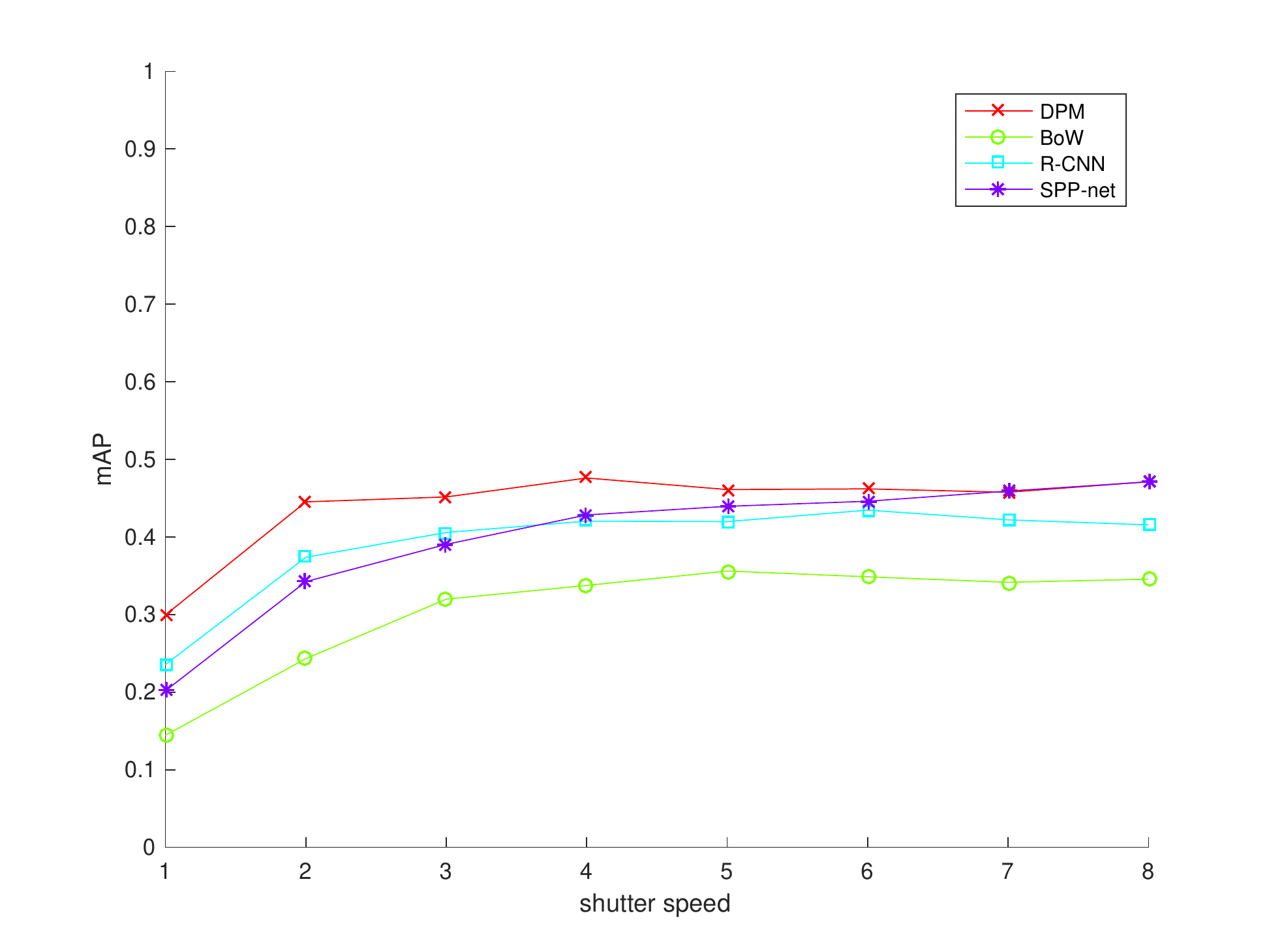}
		\caption{Shutter speed}
	\end{subfigure}
	~
	\begin{subfigure}[b]{0.23\textwidth}
		\includegraphics[draft=false,width=\textwidth]{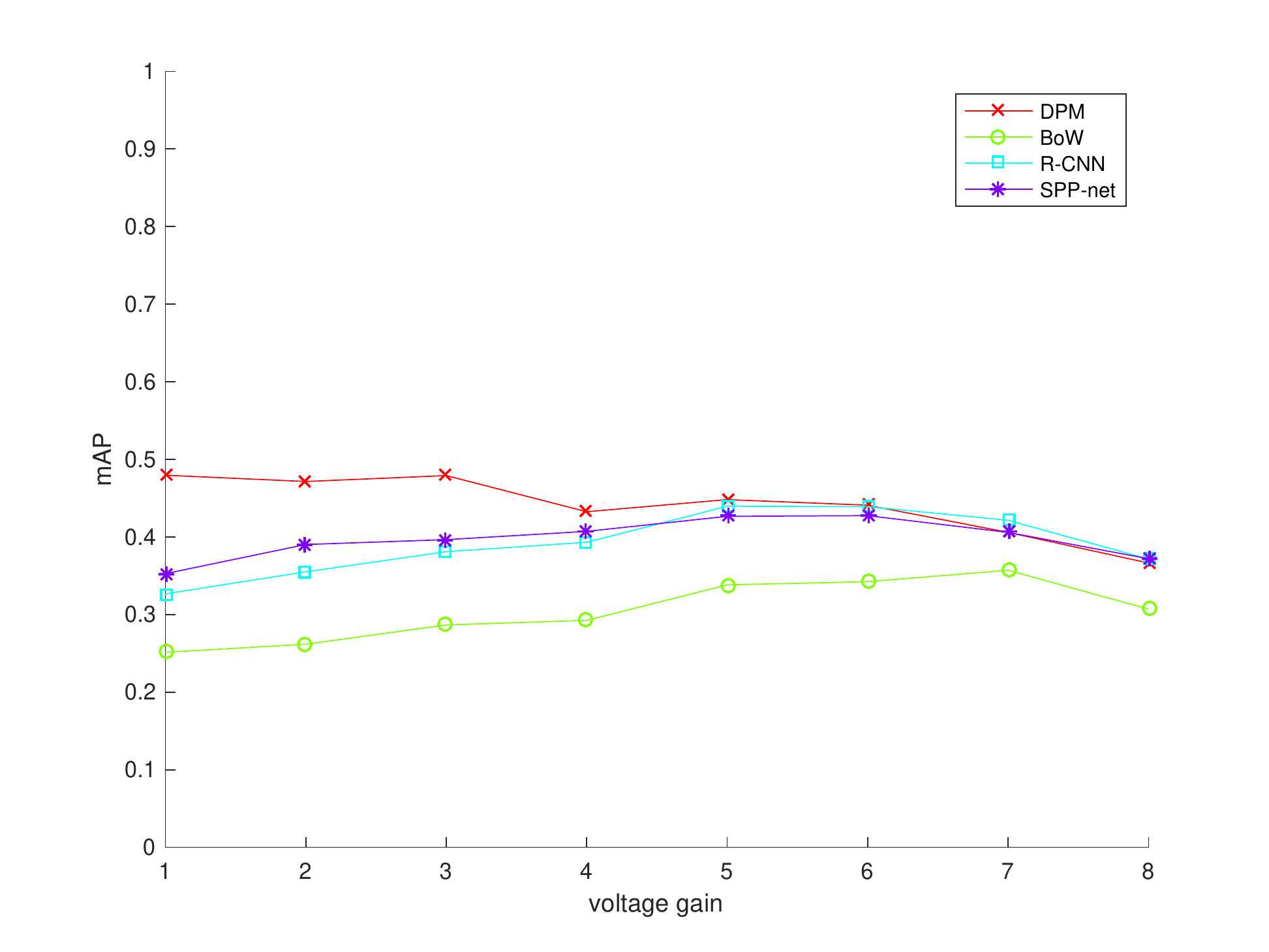}
		\caption{Voltage gain}
	\end{subfigure}
	
	\caption{The mAP of four object detection algorithms with respect to various illumination, shutter speed and voltage gain conditions.}
	\label{fig_evaluation_results_4}
\end{figure}

By aggregating the performance tables in Figure \ref{fig_evaluation_results_1} - \ref{fig_evaluation_results_3}, the results of object detection algorithms with respect to each illumination, shutter speed and voltage gain are shown in Figure \ref{fig_evaluation_results_4}.  Results are represented by mean average precision (mAP) \cite{salton1986introduction}, which is the mean of a series of AP.

For ambient illumination, the common trend is that the performance increases, reaches the peak and then decreases as the ambient illumination goes from low to high. Note that for low illumination conditions, the DPM algorithm significantly outperforms the others.  Similar pattern is also observed for shutter speed. For voltage gain, constant performance loss has been found at DPM as the voltage gain increase. One possible reason is that voltage gain introduces noises \cite{healey1994radiometric}, which affects the results. Another possibility is that these performance transients could be due to non-uniform sample representation at the given camera parameters in the original training set, and thus our method is able to uncover statistical irregularities in the training ensemble.

\section{Active Control of Camera Parameters}
\label{active_control_of_cam_params}

\begin{figure}[h]
	\centering
	\includegraphics[draft=false,width=0.5\textwidth]{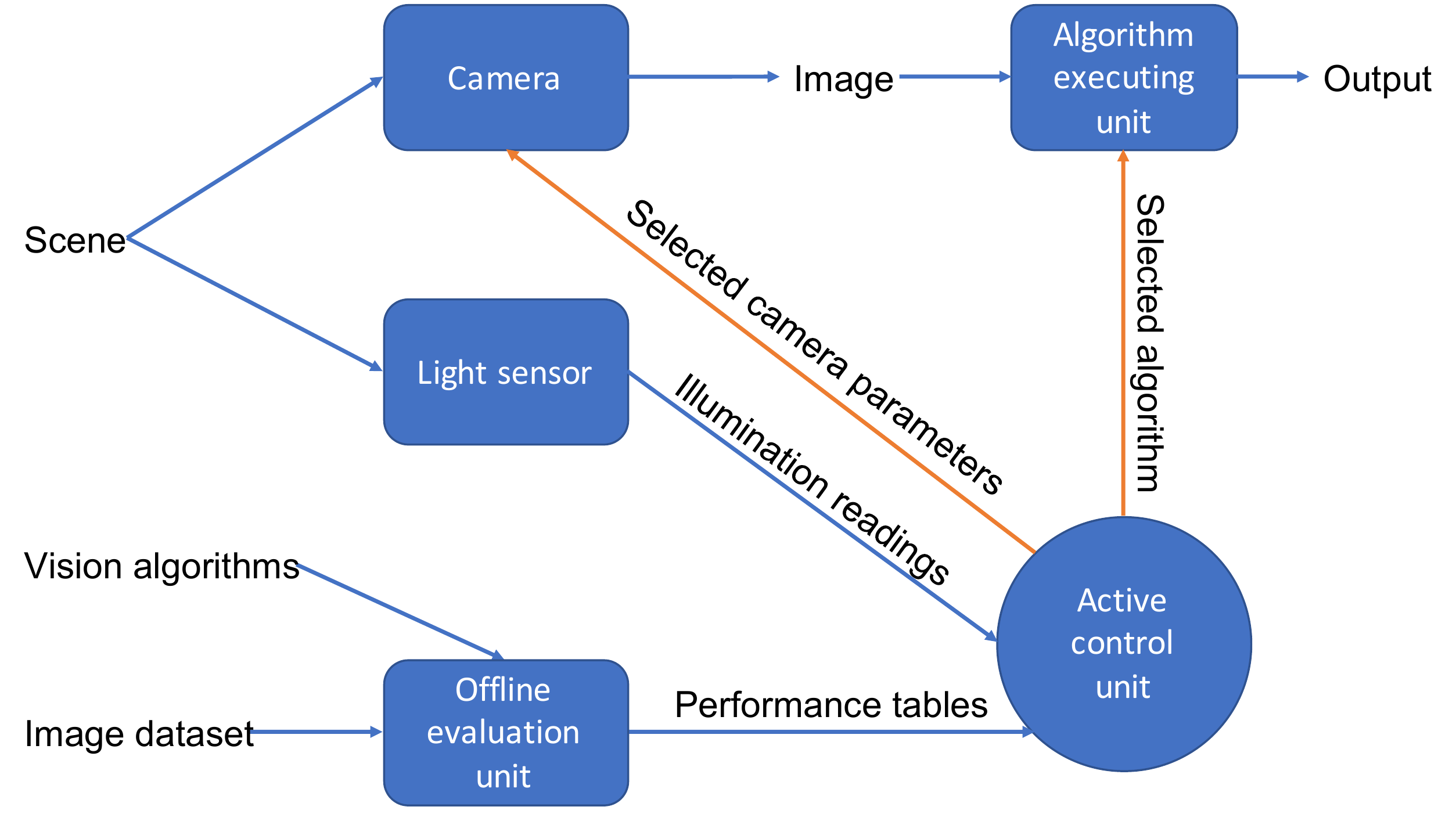}
	\caption{The proposed active control of camera parameters framework.}
	\label{fig_active_control_pipepline}
\end{figure}

As discussed in Section \ref{cam_params_on_object_detection},  the camera's intrinsic parameters have a significant impact on the performance of object detection algorithms, and the optimal shutter speed and voltage gain configurations are algorithm and ambient illumination-specific. In this section, we propose a novel active control of camera parameters method based on the evaluation results. The overall framework is shown in Figure \ref{fig_active_control_pipepline}.

There are mainly five components in this proposed framework: 1) create a dataset of images, by sampling the ambient illumination and camera parameters of interest; 2) evaluate the performance of vision algorithms on the created dataset, and build performance tables; 3) use light sensor to measure the ambient illumination; 4) select the optimal $\langle algorithm, camera\_parameters \rangle$ combination based on the performance tables, for a given illumination reading; 5) run the selected algorithm on the image taken with the selected camera parameters.

\subsection{Motivation}
The motivation of this active control of camera parameters method is mainly from the analysis of the performance behaviors of object detection algorithms. It is observed that algorithms behave differently with respect to variant illumination, shutter speed and voltage gain. We propose to systematically analyze and encode these behaviors, and to utilize these results to improve the stability and robustness of a vision system.

\subsection{Challenges}
Despite the simplicity of the idea, there are also challenging problems to solve. The first one is the reliability of the noisy performance tables, and the second one is that there may exist multiple optimal choices. 

\begin{figure}[h]
	\centering
	\includegraphics[draft=false,width=0.45\textwidth]{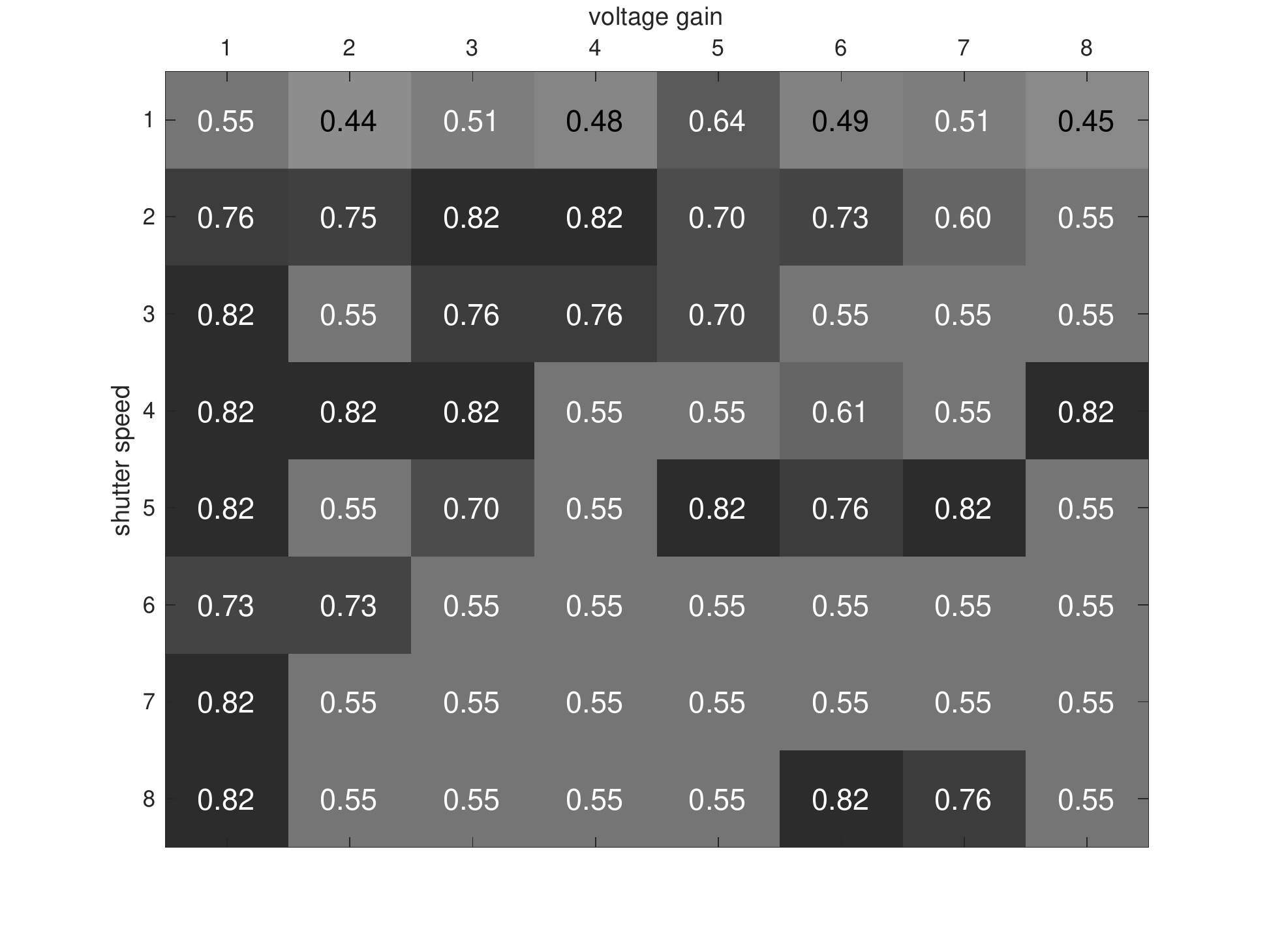}
	\caption{The performance table of DPM under the 800lx illumination condition.}
	\label{fig_challenge}
\end{figure}

Figure \ref{fig_challenge} demonstrates the original performance table of the DPM on images taken with various camera configurations, for illumination 800lx. In this case, the optimal $\langle shutter, gain \rangle$ pairs are $(2, 3)$, $(2, 4)$, $(3, 1)$, $(4, 1)$, $(4, 2)$, $(4, 3)$, $(4, 8)$, $(5, 1)$, $(5, 5)$, $(5, 7)$, $(7, 1)$, $(8, 1)$ and $(8, 6)$, which all yield the best result $0.82$. In such situation, it is unclear which one should be selected. However, it
can be found that the majority of the optimal choices are in the top-left quarter of the performance table and only a few outliers are beyond this area.

One possible reason for the outliers is that the distribution
of camera sensor parameters in the training dataset, for the
vision algorithms, is biased. The trained object detectors are
fitted to specific camera parameters combination, for various
light conditions. It may be that the unevenness of the results is due to non-uniform 
training sample distributions with respect to camera parameters and lighting conditions.
This may highlight a major difficulty with large large image datasets if their creators do not pay careful attention to 
the statistical characteristics of the training population with respect to its
key parameters.

\subsection{Implementation}
To solve the aforementioned issues, Gaussian smoothing is applied to the original performance tables.

The reason for smoothing is to remove outliers and reduce the possibility of multiple-maxima. The Guassian filter is used, due to its simplicity to trade off between each individual value and the local averages via the $\sigma$ parameter. In our implementation, the kernel size is 3 x 3 and the $\sigma$ value is configured to be 0.5, 1 or 2.  For the values at boundaries, there is not enough data to do a full smoothing operation. In such cases, we crop the Gaussian filter accordingly (zero-padding could be an alternative for the border effects). Note that the Gaussian smoothing could bring in the following side-effect on the performance tables: a low value which indicates a bad detection rate can have a higher value due to high values around it; then, this low detection value's settings would be chosen. Increase data samples or decrease the value of $\sigma$ would help avoid this situation.

For the purpose of this work, the values of ambient illumination, shutter speed and voltage gain are sampled at distinct values. However, the actual readings of light sensor and the camera's intrinsic parameters are continuous. To make the proposed system accept continuous illumination measurements and output continuous camera parameters, linear interpolation is applied.

\section{Experimental Results}
\label{experimental_results}

To evaluate how the proposed active control of camera parameters method work, empirical experiment was conducted. Our method was compared with the conventional approach, camera's built-in auto-exposure algorithm. We measured the performance of object detection algorithms using these two different approaches.

\subsection{Experimental Setup}
This experiment was conducted on the dataset introduced in Section \ref{dataset}.  We split this dataset into two groups, \textit{training} and \textit{testing}. The \textit{training} set was used for the active control of camera parameters system to compute the performance tables, and the \textit{test} set was used for testing. This process was repeated by using different combination of the \textit{training} and \textit{testing} sets, and averaging was applied to the results. The procedures were as follows:
\begin{enumerate}
	\item Pre-compute the performance tables on the \textit{training} set;
	\item For each object and for each illumination, run our proposed system as described in Section \ref{active_control_of_cam_params} to get the optimal $\langle shutter, gain\rangle$ pair;
	\item Run each object detection algorithm on the image that corresponds to the proposed camera parameters, and on the image that is taken with auto-exposure;
	\item Evaluate and compare the results (A predicted bounding box is considered correct if it overlaps no less than 50\% with the ground-truth bounding box, otherwise false).
\end{enumerate}

The auto-exposure method is the built-in exposure algorithm in Point Grey Flea3 cameras. This algorithm is controlled by two parameters, the optimal brightness level and the region-of-interest (ROI). It determines a proper exposure based the mean brightness over the ROI.  Both parameters were kept at their default values during the experiment.

For the active control method, there are also two parameters, the kernel size and the Gaussian $\sigma$.  We were using a 3x3 kernel, considering the performance table is at 8x8 resolution. We used different $\sigma$ values, i.e. 0.5, 1 and 2.

\subsection{Results and Discussions}

\begin{figure}[h]
	\centering
	\includegraphics[draft=false,width=0.36\textwidth]{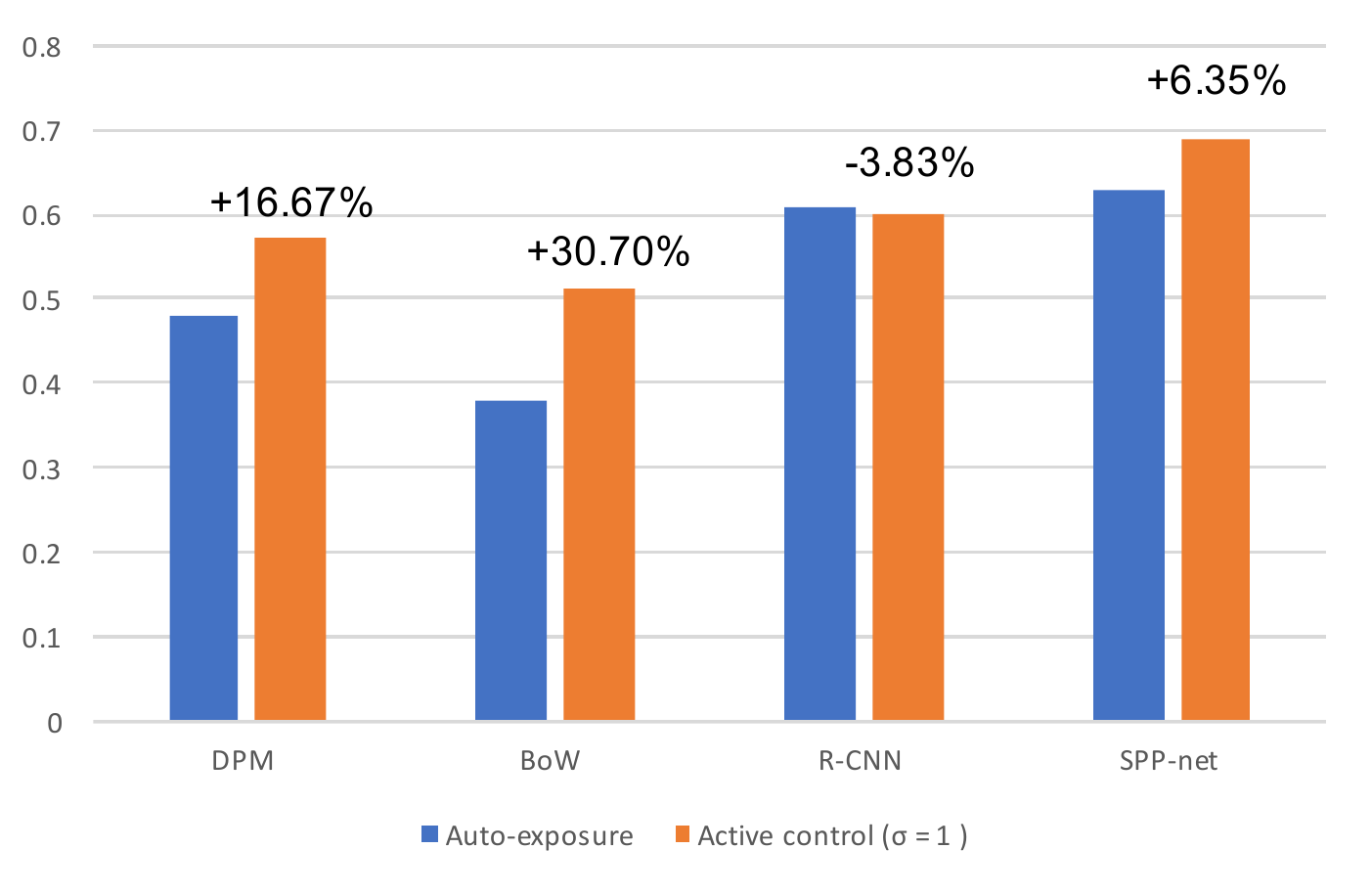}
	\caption{The comparison of auto-exposure and active control by the performance of four object detection algorithms.}
	\label{fig_compare}
\end{figure}

Table \ref{table_results} summarizes the performance of each object detection algorithm with auto-exposure and active control. Compared with auto-exposure, active control results in significantly better performance for three object detection algorithms, the DPM, Bow and R-CNN.  See Figure \ref{fig_compare} for the comparison of these two approaches by relative increments.

The performance boost is more obvious on local feature-based algorithms (DPM and Bow), than on convolutional neural network-based algorithms (R-CNN and SPP-net). One possible reason is that local features, such as SIFT and HoG, are  sensitive to the camera parameters, as pointed out in \cite{andreopoulos2012sensor}. On the contrary, convolutional  neural networks often contain a few pooling layers, which mitigate the effects of camera exposure.

Also, the results are dependent on the parameter $\sigma$ of the Gaussian smoothing operator, as noises in the performance tables of different algorithms vary. No single $\sigma$ value, that constantly outperforms the others, has been found. However, $\sigma = 1$ gives an overall decent results in our experiment.

\begin{table}
	\centering
	\begin{tabular}{| c | c | c | c | c|}
		\hline \multirow{2}{*}{} & \multirow{2}{*}{auto-exposure} & \multicolumn{3}{c|}{active control}  \\
		\cline{3-5} & & $\sigma = 0.5$ & $\sigma = 1$ & $\sigma = 2$ \\
		\hline \textit{DPM} & 0.48 & 0.51 & 0.57 & \textbf{0.60} \\
		\hline \textit{BoW} & 0.38 & 0.49 & \textbf{0.51} & 0.49 \\
		\hline \textit{R-CNN} & \textbf{0.61} & 0.56 & 0.60 & 0.60 \\
		\hline \textit{SPP-net} & 0.63 & 0.66 & \textbf{0.69} & 0.66 \\
		\hline
	\end{tabular}
	\caption{The mAP of four object detection algorithms  with auto-exposure and active control. The best performance is highlighted for each algorithm.}
	\label{table_results}
\end{table}

\section{Conclusion} 
\label{conclusion}

In this paper, a novel active control of camera parameters method is proposed in order to improve the robustness and adaptivity of vision guided robotic systems, with an emphasis on object detection algorithms. 

We first introduced a novel image dataset which incorporates ambient illumination and camera's intrinsic parameters. Then, we presented our quantitative evaluation on the performance of object detection algorithms with respect to light conditions and sensor configurations. Our results reveal the sensor bias of vision algorithms, which necessitates a finer control of camera parameters for these algorithms to work in real-world applications.

Further, we proposed the active control of camera parameters method, from the perspective of algorithm performance. This approach was empirically evaluated and compared with conventional auto-exposure method. Experimental results demonstrated the effectiveness of our proposed method in improving the robustness of vision algorithms  to illumination variation.

More importantly, our work act as a  proof of principle, on how to achieve illumination robustness and camera parameters optimization.  The methodology can be summarized as: 1) create image dataset by sampling the camera parameter space; 2) benchmark vision algorithms of interest on the dataset and compute performance statistics; 3) optimize camera parameters by the statistics.

\section{Future Work} 
\label{future_work}
In the experiments, the performance tables are obtained with five objects in a simple scene. In the future, the tests should be extended to include multiple objects in different backgrounds and from different viewpoints. Also, an experiment with more than one camera would demonstrate the robustness of the method.

\bibliographystyle{plainnat}
\bibliography{paper}

\begin{thebibliography}{34}
\providecommand{\natexlab}[1]{#1}
\providecommand{\url}[1]{\texttt{#1}}
\expandafter\ifx\csname urlstyle\endcsname\relax
  \providecommand{\doi}[1]{doi: #1}\else
  \providecommand{\doi}{doi: \begingroup \urlstyle{rm}\Url}\fi

\bibitem[Adini et~al.(1997)Adini, Moses, and Ullman]{adini1997face}
Yael Adini, Yael Moses, and Shimon Ullman.
\newblock Face recognition: The problem of compensating for changes in
  illumination direction.
\newblock \emph{IEEE Transactions on pattern analysis and machine
  intelligence}, 19\penalty0 (7):\penalty0 721--732, 1997.

\bibitem[Aloimonos et~al.(1988)Aloimonos, Weiss, and
  Bandyopadhyay]{aloimonos1988active}
John Aloimonos, Isaac Weiss, and Amit Bandyopadhyay.
\newblock Active vision.
\newblock \emph{International Journal of Computer Vision}, 1\penalty0
  (4):\penalty0 333--356, 1988.

\bibitem[Andreopoulos and Tsotsos(2012)]{andreopoulos2012sensor}
Alexander Andreopoulos and John~K Tsotsos.
\newblock On sensor bias in experimental methods for comparing interest-point,
  saliency, and recognition algorithms.
\newblock \emph{IEEE Transactions on Pattern Analysis and Machine
  Intelligence}, 34\penalty0 (1):\penalty0 110--126, 2012.

\bibitem[Bajcsy(1985)]{bajcsy1985active}
Ruzena Bajcsy.
\newblock Active perception vs. passive perception.
\newblock In \emph{IEEE Workshop on Computer Vision Representation and
  Control}, Bellaire, Michigan, 1985.

\bibitem[Belhumeur and Kriegman(1998)]{belhumeur1998set}
Peter~N Belhumeur and David~J Kriegman.
\newblock What is the set of images of an object under all possible
  illumination conditions?
\newblock \emph{International Journal of Computer Vision}, 28\penalty0
  (3):\penalty0 245--260, 1998.

\bibitem[Bileschi()]{bileschi2007cbcl}
Stanely Bileschi.
\newblock {CBCL} streetscenes challenge framework (2007).
\newblock http://cbcl.mit.edu/software-datasets/streetscenes/.

\bibitem[Brunnstr{\"o}m et~al.(1996)Brunnstr{\"o}m, Eklundh, and
  Uhlin]{brunnstrom1996active}
Kjell Brunnstr{\"o}m, Jan-Olof Eklundh, and Tomas Uhlin.
\newblock Active fixation for scene exploration.
\newblock \emph{International Journal of Computer Vision}, 17\penalty0
  (2):\penalty0 137--162, 1996.

\bibitem[Dalal and Triggs(2005)]{dalal2005histograms}
Navneet Dalal and Bill Triggs.
\newblock Histograms of oriented gradients for human detection.
\newblock In \emph{IEEE Conference on Computer Vision and Pattern Recognition},
  volume~1, pages 886--893. IEEE, 2005.

\bibitem[Everingham et~al.(2010)Everingham, Van~Gool, Williams, Winn, and
  Zisserman]{everingham2010pascal}
Mark Everingham, Luc Van~Gool, Christopher~KI Williams, John Winn, and Andrew
  Zisserman.
\newblock The {PASCAL} visual object classes ({VOC}) challenge.
\newblock \emph{International Journal of Computer Vision}, 88\penalty0
  (2):\penalty0 303--338, 2010.

\bibitem[Felzenszwalb et~al.(2010)Felzenszwalb, Girshick, McAllester, and
  Ramanan]{felzenszwalb2010object}
Pedro~F Felzenszwalb, Ross~B Girshick, David McAllester, and Deva Ramanan.
\newblock Object detection with discriminatively trained part-based models.
\newblock \emph{IEEE Transactions on Pattern Analysis and Machine
  Intelligence}, 32\penalty0 (9):\penalty0 1627--1645, 2010.

\bibitem[Girshick et~al.(2014)Girshick, Donahue, Darrell, and
  Malik]{girshick2014rich}
Ross Girshick, Jeff Donahue, Trevor Darrell, and Jagannath Malik.
\newblock Rich feature hierarchies for accurate object detection and semantic
  segmentation.
\newblock In \emph{Proc. IEEE Conf. on Computer Vision and Pattern
  Recognition}, pages 580--587. IEEE, 2014.

\bibitem[Han et~al.(2013)Han, Shan, Chen, and Gao]{han2013comparative}
Hu~Han, Shiguang Shan, Xilin Chen, and Wen Gao.
\newblock A comparative study on illumination preprocessing in face
  recognition.
\newblock \emph{Pattern Recognition}, 46\penalty0 (6):\penalty0 1691--1699,
  2013.

\bibitem[He et~al.(2015)He, Zhang, Ren, and Sun]{he2015spatial}
Kaiming He, Xiangyu Zhang, Shaoqing Ren, and Jian Sun.
\newblock Spatial pyramid pooling in deep convolutional networks for visual
  recognition.
\newblock \emph{IEEE Transactions on Pattern Analysis and Machine
  Intelligence}, 37\penalty0 (9):\penalty0 1904--1916, 2015.

\bibitem[Healey and Kondepudy(1994)]{healey1994radiometric}
Glenn~E Healey and Raghava Kondepudy.
\newblock Radiometric {CCD} camera calibration and noise estimation.
\newblock \emph{IEEE Transactions on Pattern Analysis and Machine
  Intelligence}, 16\penalty0 (3):\penalty0 267--276, 1994.

\bibitem[Johnson(1984)]{johnson1984photographic}
Bruce~K Johnson.
\newblock Photographic exposure control system and method, January~3 1984.
\newblock US Patent 4,423,936.

\bibitem[Lin et~al.(2014)Lin, Maire, Belongie, Hays, Perona, Ramanan,
  Doll{\'a}r, and Zitnick]{lin2014microsoft}
Tsung-Yi Lin, Michael Maire, Serge Belongie, James Hays, Pietro Perona, Deva
  Ramanan, Piotr Doll{\'a}r, and C~Lawrence Zitnick.
\newblock Microsoft {COCO}: Common objects in context.
\newblock In \emph{European Conference on Computer Vision}, pages 740--755.
  Springer, 2014.

\bibitem[Linderoth et~al.(2013)Linderoth, Robertsson, and
  Johansson]{linderoth2013color}
Magnus Linderoth, Anders Robertsson, and Rolf Johansson.
\newblock Color-based detection robust to varying illumination spectrum.
\newblock In \emph{IEEE Workshop on Robot Vision}, pages 120--125. IEEE, 2013.

\bibitem[Llano et~al.(2006)Llano, Vazquez, Kittler, and
  Messer]{llano2006illumination}
Eduardo~Garea Llano, Heydi~Mendez Vazquez, Josef Kittler, and Kieron Messer.
\newblock An illumination insensitive representation for face verification in
  the frequency domain.
\newblock In \emph{International Conference on Pattern Recognition}, volume~1,
  pages 215--218. IEEE, 2006.

\bibitem[Lowe(2004)]{lowe2004distinctive}
David~G Lowe.
\newblock Distinctive image features from scale-invariant keypoints.
\newblock \emph{International journal of computer vision}, 60\penalty0
  (2):\penalty0 91--110, 2004.

\bibitem[Lu et~al.(2010)Lu, Zhang, Yang, and Zheng]{lu2010camera}
Huimin Lu, Hui Zhang, Shaowu Yang, and Zhiqiang Zheng.
\newblock Camera parameters auto-adjusting technique for robust robot vision.
\newblock In \emph{IEEE International Conference on Robotics and Automation},
  pages 1518--1523. IEEE, 2010.

\bibitem[Maier et~al.(2011)Maier, Eschey, and Steinbach]{maier2011image}
Werner Maier, Michael Eschey, and Eckehard Steinbach.
\newblock Image-based object detection under varying illumination in
  environments with specular surfaces.
\newblock In \emph{IEEE International Conference on Image Processing}, pages
  1389--1392. IEEE, 2011.

\bibitem[Osadchy and Keren(2001)]{osadchy2001image}
Margarita Osadchy and Daniel Keren.
\newblock Image detection under varying illumination and pose.
\newblock In \emph{IEEE Conference on Computer Vision}, volume~2, pages
  668--673. IEEE, 2001.

\bibitem[Osadchy and Keren(2004)]{osadchy2004efficient}
Margarita Osadchy and Daniel Keren.
\newblock Efficient detection under varying illumination conditions and image
  plane rotations.
\newblock \emph{Computer Vision and Image Understanding}, 93\penalty0
  (3):\penalty0 245--259, 2004.

\bibitem[Russakovsky et~al.(2015)Russakovsky, Deng, Su, Krause, Satheesh, Ma,
  Huang, Karpathy, Khosla, Bernstein, et~al.]{russakovsky2015imagenet}
Olga Russakovsky, Jia Deng, Hao Su, Jonathan Krause, Sanjeev Satheesh, Sean Ma,
  Zhiheng Huang, Andrej Karpathy, Aditya Khosla, Michael Bernstein, et~al.
\newblock Imagenet large scale visual recognition challenge.
\newblock \emph{International Journal of Computer Vision}, 115\penalty0
  (3):\penalty0 211--252, 2015.

\bibitem[Salton and McGill(1986)]{salton1986introduction}
Gerard Salton and Michael~J McGill.
\newblock Introduction to modern information retrieval.
\newblock 1986.

\bibitem[Sampat et~al.(1999)Sampat, Venkataraman, Yeh, and
  Kremens]{sampat1999system}
Nitin Sampat, Shyam Venkataraman, Thomas Yeh, and Robert~L Kremens.
\newblock System implications of implementing auto-exposure on consumer digital
  cameras.
\newblock In \emph{Electronic Imaging'99}, pages 100--107. International
  Society for Optics and Photonics, 1999.

\bibitem[Shim et~al.(2014)Shim, Lee, and Kweon]{shim2014auto}
Inwook Shim, Joon-Young Lee, and In~So Kweon.
\newblock Auto-adjusting camera exposure for outdoor robotics using gradient
  information.
\newblock In \emph{IEEE/RSJ International Conference on Intelligent Robots and
  Systems}, pages 1011--1017. IEEE, 2014.

\bibitem[Tanaka et~al.(2009)Tanaka, Shimada, Arita, and
  Taniguchi]{tanaka2009object}
Tatsuya Tanaka, Atsushi Shimada, Daisaku Arita, and Rin-ichiro Taniguchi.
\newblock Object detection under varying illumination based on adaptive
  background modeling considering spatial locality.
\newblock In \emph{Pacific-Rim Symposium on Image and Video Technology}, pages
  645--656. Springer, 2009.

\bibitem[Tang et~al.(2012)Tang, Salakhutdinov, and Hinton]{TangSH12b}
Yichuan Tang, Ruslan Salakhutdinov, and Geoffrey Hinton.
\newblock Deep lambertian networks.
\newblock In \emph{International Conference on Machine Learning}, 2012.

\bibitem[Tenenbaum(1970)]{tenenbaum1970accommodation}
Jay~Martin Tenenbaum.
\newblock Accommodation in computer vision.
\newblock Technical report, DTIC Document, 1970.

\bibitem[Tsotsos(1992)]{tsotsos1992relative}
John~K Tsotsos.
\newblock On the relative complexity of active vs. passive visual search.
\newblock \emph{International Journal of Computer Vision}, 7\penalty0
  (2):\penalty0 127--141, 1992.

\bibitem[Uijlings et~al.(2013)Uijlings, van~de Sande, Gevers, and
  Smeulders]{uijlings2013selective}
Jasper~RR Uijlings, Koen~EA van~de Sande, Theo Gevers, and Arnold~WM Smeulders.
\newblock Selective search for object recognition.
\newblock \emph{International Journal of Computer Vision}, 104\penalty0
  (2):\penalty0 154--171, 2013.

\bibitem[Wei and Lai(2004)]{wei2004robust}
Shou-Der Wei and Shang-Hong Lai.
\newblock Robust face recognition under lighting variations.
\newblock In \emph{International Conference on Pattern Recognition}, volume~1,
  pages 354--357. IEEE, 2004.

\bibitem[Westerhoff et~al.(2015)Westerhoff, Meuter, and
  Kummert]{westerhoff2015generic}
Jens Westerhoff, Mirko Meuter, and Anton Kummert.
\newblock A generic parameter optimization workflow for camera control
  algorithms.
\newblock In \emph{IEEE International Conference on Intelligent Transportation
  Systems}, pages 944--949. IEEE, 2015.

\end{thebibliography}

\end{document}